%% file: acl_latex.tex
\documentclass[11pt]{article}

\usepackage[final]{acl}
\usepackage{times}
\usepackage{latexsym}
\usepackage{xcolor}
\usepackage{booktabs}
\definecolor{forestgreen}{RGB}{34, 139, 34}      
\usepackage{tcolorbox}
\usepackage[T1]{fontenc}

\usepackage[utf8]{inputenc}

\usepackage{microtype}

\usepackage{inconsolata}

\usepackage{graphicx}
\usepackage{amsmath}
\usepackage{booktabs}

\usepackage{booktabs}

\usepackage[table]{xcolor}

\usepackage{arydshln}

\definecolor{headerblue}{RGB}{30, 55, 100}

\definecolor{comprow}{RGB}{240, 240, 245}
\title{Can LLM Annotators Match Human Consistency? Benchmarking Human Agreement and Optimizing Multi-Agent with Contrastive Knowledge}

\author{Xiaopeng Ke$^{1,2}$~~
        Zhixin Wang$^1$~~~~
        Xuebo Liu$^2$\thanks{~Xuebo Liu is the corresponding author.}~~
        Xinyu Ma$^2$~~
        Xing Zheng$^1$~~\\
        \textbf{Bowen Song}$^1$~~
        \textbf{Weiqiang Wang}$^1$~~
        \textbf{Min Zhang}$^2$~~\\
        $^1$ANT Group~~~~
        $^2$Harbin Institute of Technology (Shenzhen) \\ 
        \texttt{\{xiaopk7,mxinyuma\}@gmail.com}\;
        \texttt{\{liuxuebo,zhangmin2021\}@hit.edu.cn} \\
        \texttt{\{linglan.wzx, feishang.zx, bowen.sbw, weiqiang.wwg\}@antgroup.com}
}
\begin{document}
\maketitle


\begin{abstract}
LLM-based annotation has advanced rapidly, yet existing benchmarks remain misaligned with real-world practice. Actual annotation often involves lengthy, task-specific guidelines, and different annotators may produce divergent but reasonable judgments, especially on boundary cases, where single-labeled dataset fail to reflect this disagreement. This gap also limits existing single-agent and workflow-based annotation systems, which often lack adaptability to complex, evolving guidelines. To bridge this gap, we introduce \textbf{ANT-A Benchmark}, a real-world multi-annotator benchmark spanning 7 datasets across domains and modalities, with task-specific guidelines and 5 independent human labels per instance. ANT-A enables a distribution-aware evaluation, using Fleiss’ $\kappa$ to assess how closely model annotations align with human labeling distributions. We further propose \textbf{ANT-OR}, a multi-agent annotation framework that refines prompts by distilling contrastive decision rules from positive and negative feedback. Experiments show that ANT-OR improves agreement@$\ge$2 accuracy by 4.5 points over the strongest baseline and achieves the highest alignment with human annotator distributions as measured by Fleiss' $\kappa$ within 2 points. Code and benchmark are available at~\url{https://github.com/Krueske/ANT-OR}.


\end{abstract}


\section{Introduction}
\label{sec:intro}

Data annotation is essential to machine learning, yet acquiring high-quality annotations at scale remains costly and challenging~\cite{2017Snorkel,zhang2021wrench,10.1145/3613904.3641960}. Large language models (LLMs) have emerged as an attractive option for automated annotation across diverse tasks~\cite{he-etal-2024-annollm,tan-etal-2024-large}. However, current LLM-based annotation systems still depend heavily on hand-crafted prompts and static workflows~\cite{yu2025infiagentselfevolvingpyramidagent}, which often fail on complex tasks with detailed guidelines and boundary cases.

\begin{figure}[t]
  \includegraphics[width=\columnwidth]{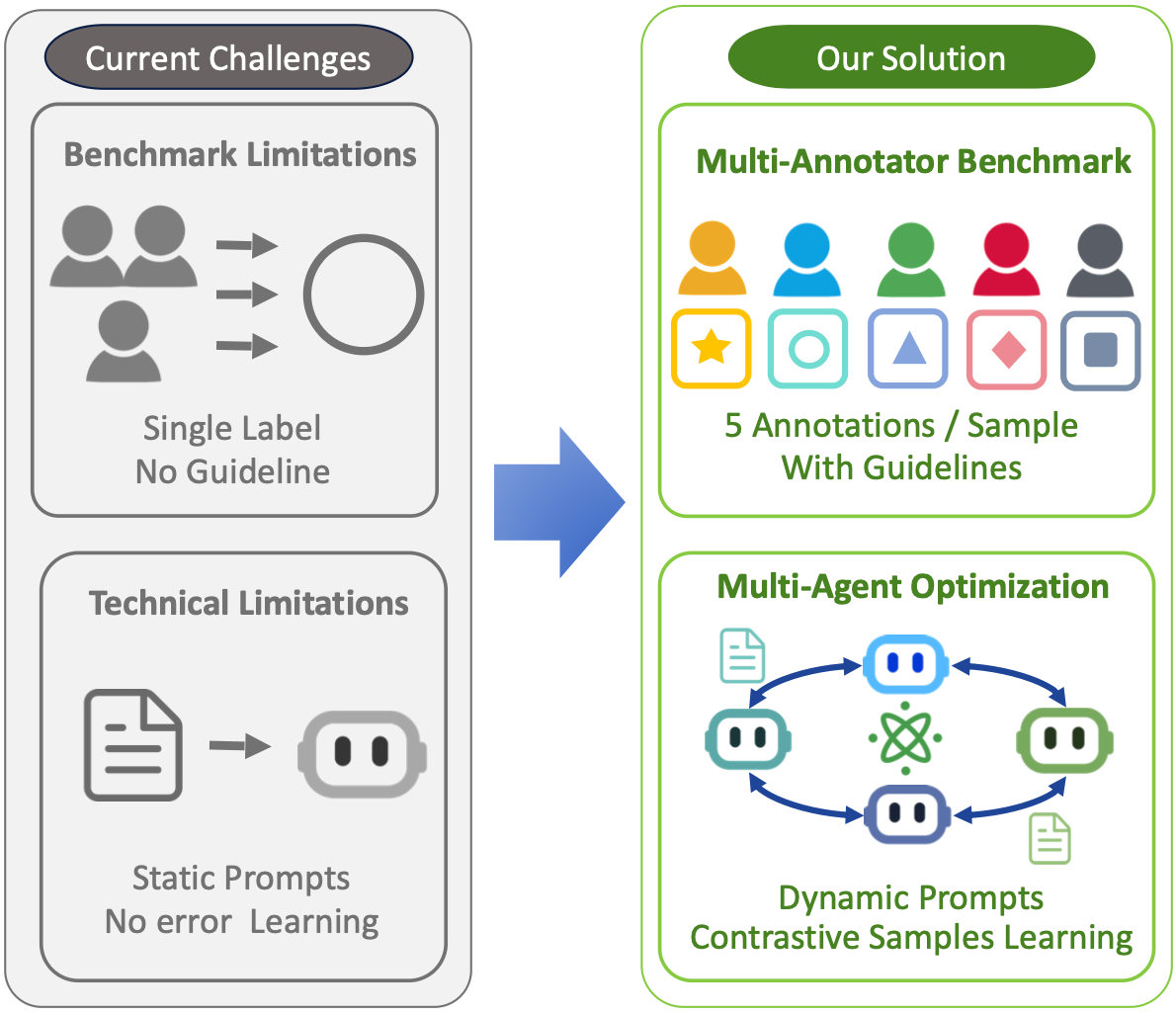}
  \caption{Overview of our contributions. We construct \textbf{ANT-A}, a benchmark with detailed guidelines and multiple independent human annotations, and develop \textbf{ANT-OR}, a multi-agent framework that improves automated annotation through iterative optimization with contrastive knowledge.}
  \label{fig:firstfigure}
\end{figure}

A major obstacle to further developing of data annotation methods is the lack of evaluation protocols that reflect real-world annotation practice. Existing LLM annotation benchmarks often lack the guideline documents used in real-world tasks and provide only a single gold-standard label per sample~\cite{basile2021we, kim2026perfectapiscomprehensiveevaluation}. However, many realistic annotation tasks are governed by detailed annotation guidelines, which can be inherently ambiguous or open to interpretation. As a result, disagreement among annotators may reflect valid alternative interpretations rather than annotation errors~\cite{tseng2025evaluating, fleisig-etal-2023-majority}. 
Single-label benchmarks thus collapse this uncertainty and cannot determine whether LLM annotators achieve human-level consistency, e.g., Fleiss' $\kappa$~\cite{fleiss1971kappa}, or simply match one reference answer~\cite{mehrotra2026multiperspectivellmannotationsvalid}.
To address this gap, we construct \textbf{ANT-A} (\textbf{A}ggregated \textbf{N}-annotator \textbf{T}ask \textbf{A}nnotations), a multi-annotator benchmark comprising 7 datasets with detailed guidelines and 5 independent human annotations per sample.


Realistic annotation settings also raise a challenge for single-agent systems, which often struggle on complex tasks governed by guidelines because a single reasoning perspective cannot reliably handle nuanced criteria and edge cases~\cite{ban2026completing,pan2025why}. Multi-agent systems offer a promising alternative through complementary judgments and cross-checking~\cite{tran2025multiagentcollaborationmechanismssurvey}. Existing approaches typically treat agents as static components~\cite{Zheng2025LifelongLO} and fail to transform annotation experience into structured knowledge~\cite{abdollahi2025demystifyingerrorsllmreasoning}.
To address this limitation, we propose \textbf{ANT-OR}, an automated \textbf{AN}notation framework \textbf{T}hrough multi-agent \textbf{O}ptimization with agent-level cont\textbf{R}astive knowledge. By incorporating agent-level contrastive knowledge into optimization, ANT-OR updates decision heuristics to improve future judgments and enhance inter-agent coordination.


Extensive experiments on the proposed benchmark show that ANT-OR substantially improves annotation performance across diverse tasks, outperforming the strongest baseline by 4.5 points in accuracy and achieves the highest agreement with human annotator distributions as measured by Fleiss' $\kappa$ within 2 points on average. Our primary contributions are summarized as follows:
\begin{itemize}
    \item We identify two critical bottlenecks to reliable automated annotation: inadequate benchmarks that fail to capture inter-annotator disagreement and complexity from detailed annotation guidelines, and limitations of current annotation agents, particularly those based on single-agent or static multi-agent designs.
    \item We construct and release a multi-annotator benchmark comprising 7 datasets with detailed guidelines and 5 independent human annotations per sample, enabling a distribution-aware evaluation, using Fleiss’ $\kappa$ to assess how closely model annotations align with human labeling distributions.
    \item We propose ANT-OR, an adaptive multi-agent framework that can transform collective prediction successes and failures into reusable structured decision knowledge, enabling more robust collaboration on complex annotation tasks involving nuanced decision boundaries.
\end{itemize}


\section{Related Work}
\label{sec:related}
\paragraph{LLM-based Data Annotation and Benchmarks.}
LLMs are increasingly used as scalable alternatives to human annotators~\cite{he-etal-2024-annollm,tan-etal-2024-large,gilardi2023chatgpt}, creating a need for reliable benchmarks of their annotation behavior. 
However, existing benchmarks often simplify real-world annotation in two ways: first, they lack the detailed guideline documents that capture the complexity of realistic annotation tasks; second, they provide only a single gold or aggregated label per sample~\cite{basile2021we,kim2026perfectapiscomprehensiveevaluation}. 
These limitations make it difficult to evaluate whether LLMs can follow task-specific annotation protocols~\cite{ke-etal-2025-aquilt, liu-etal-2024-selectit} and whether their outputs match the distribution of human judgments rather than a single reference answer~\cite{liu-etal-2024-uncertainty}. 
Recent work shows that LLMs struggle with specialized guidelines~\cite{tseng2025evaluating,diao2025guidebench}, and that annotator disagreement often reflects meaningful differences in perspective rather than noise~\cite{santurkar2024whose,zhang2025diverging,chung2025bridging,wang2026humanefficientaligningannotations}. 
Concurrently, a line of work models annotator disagreement as a first-class problem rather than noise to be averaged out. 
\citet{fleisig2023majority} and \citet{xu2025modeling} predict individual annotator labels or demographic-group distributions. 
In emotion recognition, \citet{wu2026amber} and \citet{jia2026decoding} model ambiguity directly from human label distributions without demographic routing, yet their approaches are tied to task-specific architectures. 
These methods are valuable references, but they do not address the cross-domain, guideline-grounded setting we target, where disagreement persists on boundary cases even under explicit rules. 
Overall, prior work recognizes the importance of both guidelines and disagreement, but existing benchmarks typically study them in isolation. 
We address this gap with \textbf{ANT-A}, a multi-annotator benchmark that pairs detailed annotation guidelines with multiple independent human annotations per sample.

\begin{figure*}[t]
  \centering
  \includegraphics[width=\textwidth]{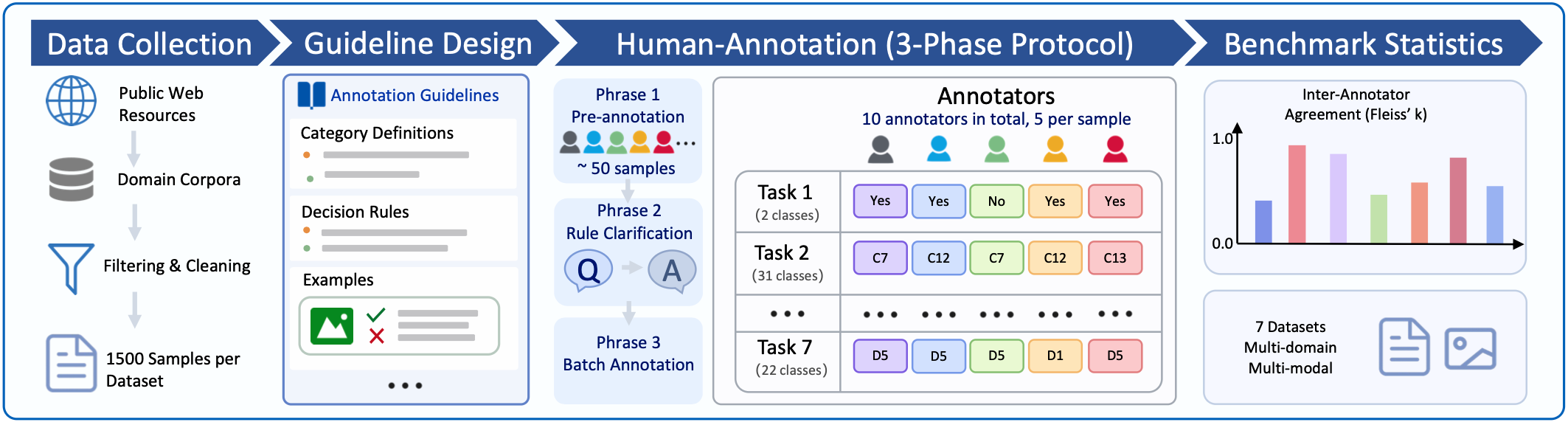}
  \caption{Construction of \textbf{ANT-A}. We curate domain-relevant samples, establish annotation guidelines through a pre-annotation pilot, and collect 5 independent annotations per sample to measure inter-annotator agreement.}
  \label{fig:benchmark_construction}
\end{figure*}

\paragraph{Multi-Agent System Construction and Optimization.}
Multi-agent collaboration and optimization have emerged as complementary directions for improving LLM reliability. Existing approaches refine outputs through sampling~\cite{wang2023selfconsistency}, inter-agent dialogue~\cite{du2024agentdebate,liang2024encouraging}, or generation-verification pipelines~\cite{hegazy2025mafa,ahtisham2025orchestration}, but treat prompts as static after initial design. Recent work explores optimizing multi-agent structures via sparse topology learning~\cite{zhang2025cut,zhang2025gdesigner, wang-etal-2025-agentdropout, tian-etal-2025-agentinit} or joint prompt-topology search~\cite{zhou2025mass, wang-etal-2026-maspo}, yet these target token efficiency rather than task-specific adaptation with cross-iteration transfer. Meanwhile, evolutionary~\cite{guo2024connecting,fernando2024promptbreeder}, trajectory-based~\cite{yang2024large,tang2024gpo}, and feedback-based methods~\cite{pryzant2023automatic,xiang2025self,chen-etal-2025-sgic} optimize prompts without fine-grained error analysis or cross-iteration knowledge retention. Memory-based ERM~\cite{yan-etal-2025-efficient} accumulates past errors but operates at the instance level, lacking structural abstraction of error patterns. At the model-training level, weakness-case acquisition has also been used to focus iterative preference optimization on model-specific failures \citep{rao-etal-2025-apt, rao-etal-2025-seapo}. ANT-OR instead mines structured decision rules from iterative error-success contrasts, enabling multi-agent systems to rapidly adapt to complex annotation tasks without training.

\begin{table}[t]
\centering
\scalebox{0.48}{
\begin{tabular}{lccccc}
\toprule
\textbf{Dataset} & \textbf{Domain} & \textbf{Modality} & \textbf{\#Samples} & \textbf{\#Classes} & \textbf{Fleiss' $\kappa$} \\
\midrule
Disease Privacy Assessment & Medical & Text & 1,500 & 2 & 38.77 \\
Billing Scenario Classification & E-commerce & Text & 1,500 & 31 & 83.50 \\
Medical Entity Extraction & Medical & Text & 1,500 & 22 & 82.09 \\
Social Text Classification & General & Text & 1,500 & 15 & 44.62 \\
Image Attribution Recognition & General & Vision-Language & 1,500 & 3 & 58.59 \\
Image Classification & E-commerce & Vision-Language & 1,500 & 24 & 81.27 \\
Image Text Relevance & General & Vision-Language & 1,500 & 3 & 56.37 \\
\bottomrule
\end{tabular}
}
\caption{Summary statistics of ANT-A}
\label{tab:benchmark_stats}
\end{table}

\section{ANT-A Benchmark}
\label{sec:benchmark}

As shown in Figure~\ref{fig:benchmark_construction}, we introduce \textbf{ANT-A} (\textbf{A}ggregated \textbf{N}-annotator \textbf{T}ask \textbf{A}nnotations), a multi-source benchmark that jointly provides detailed guidelines and 5 independent human judgments per sample that reflect the inherent ambiguity of tasks and subjectivity of human annotations rendering single labels unreliable (see Appendix~\ref{sec:case_study}), enabling rigorous comparison between LLM systems and human raters under realistic conditions. 

\subsection{Dataset Preparation}
\label{sec:dataset_preparation}
\paragraph{Data Collection.} 
We collect domain-relevant query samples from public internet data sources for each dataset, anonymize potentially identifiable information, and filter exact duplicates and low-quality instances to obtain 1,500 samples with a mix of clear-cut and boundary cases.
\paragraph{Guideline Document Design.} 
We derive annotation guidelines from real-world requirements and refine them by clarifying ambiguous definitions, adding decision rules for boundary cases, and tailoring to our data distribution. The resulting documents specify fine-grained decision boundaries and edge-case handling rules.

\begin{figure*}[t]
  \centering
  \includegraphics[width=\textwidth]{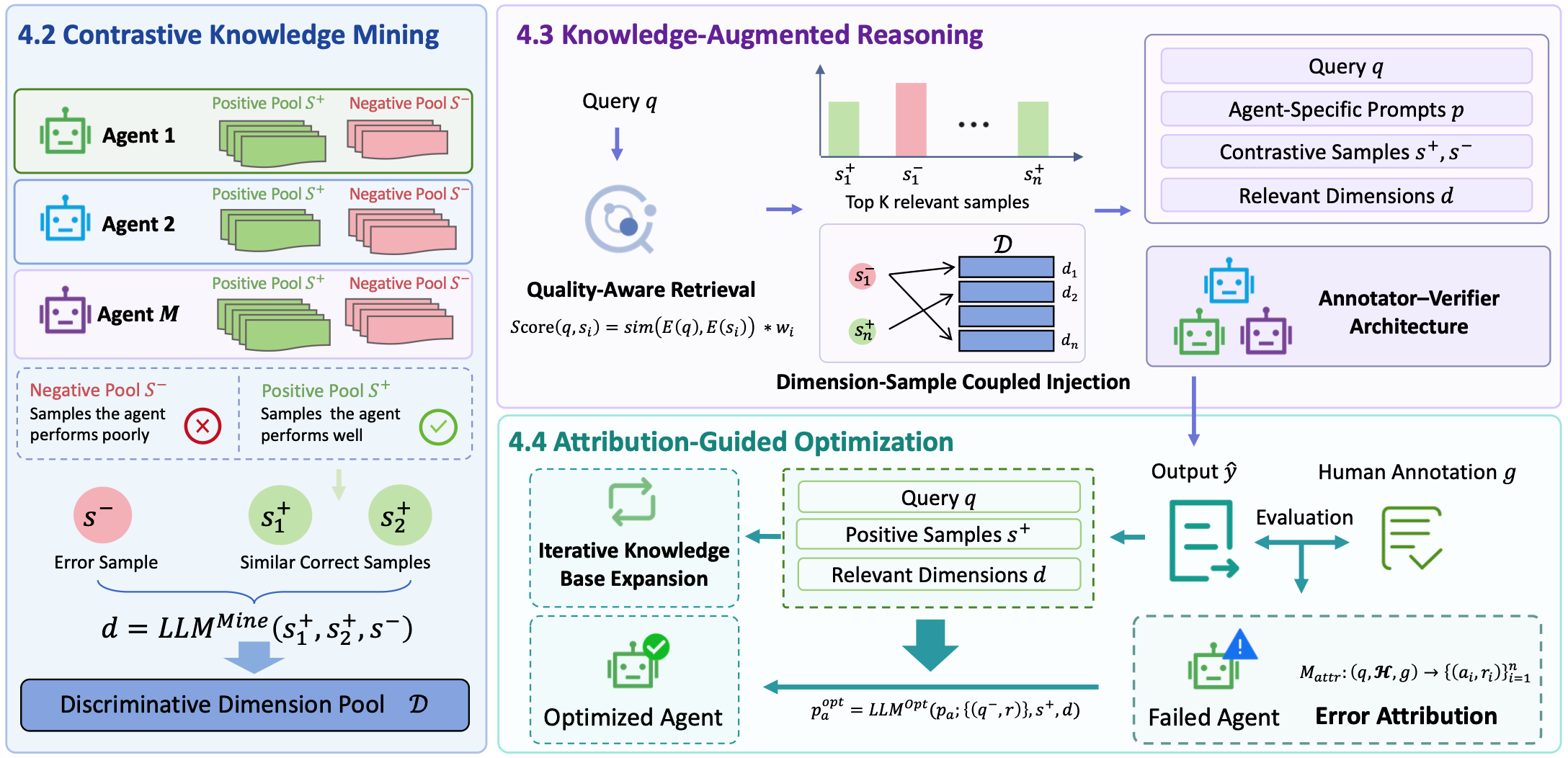}
  \caption{
  Workflow of \textbf{ANT-OR}. Agents construct contrastive knowledge bases (\S\ref{sec:knowledge_base_construction}), perform retrieval-augmented annotation (\S\ref{sec:mas}), and optimize prompts via failure attribution with iterative knowledge updates (\S\ref{sec:optimization}).
}
  \label{fig:framework}
\end{figure*}

\subsection{Human Annotation Process}
\label{sec:annotation_process}

All annotations are conducted by professional annotators from an internal annotation vendor, and we present the relevant backgrounds of the annotators in Table~\ref{tab:annotator_info}. We use a pool of 10 annotators, and each sample is independently labeled by 5 of them. This setup reflects realistic annotation practice and reduces bias from a fixed small human group. To ensure quality, we adopt a three-phase protocol, with a total annotation cost of USD \$4078.

\paragraph{Phase 1: Pre-annotation.}
A small subset of each dataset (approximately 50 samples) is first annotated by all 10 annotators. This pilot phase familiarizes annotators with the guideline document and task requirements, and surfaces ambiguities or under-specified rules in the guidelines. Outputs from this phase are reviewed to identify systematic disagreements stemming from guideline ambiguity rather than genuine judgment differences.

\paragraph{Phase 2: Rule Clarification.}
Based on the pre-annotation results, we collect questions and confusion points raised by annotators regarding category definitions, boundary cases, and decision rules. These are addressed through a structured Q\&A process: annotator questions are compiled, unified answers are provided, and the guideline document is revised accordingly to eliminate identified ambiguities. This iterative clarification ensures that remaining disagreements in the formal annotation phase reflect genuine task difficulty rather than inconsistent rule interpretation.

\paragraph{Phase 3: Batch Annotation.}
After guideline finalization, each of the 1,500 samples in a dataset is independently labeled by 5 annotators selected from the pool of 10. Annotators label only their assigned samples and cannot access others' labels, enabling meaningful inter-annotator agreement analysis and comparison with LLM-based annotators.

\subsection{Benchmark Statistics}
\label{sec:benchmark_stats}

Table~\ref{tab:benchmark_stats} summarizes the key statistics of our benchmark. Across the 7 datasets, we observe varying levels of inter-annotator agreement as measured by Fleiss' $\kappa$~\cite{fleiss1971kappa}, reflecting the inherent difficulty differences across tasks and domains. Across datasets, instances with five completely different annotations are extremely rare (nearly 1\%), suggesting that most samples retain partial human consensus. Further details on dataset construction, annotation guidelines, and annotator demographics are provided in Appendix~\ref{sec:benchmark_details}.

\input{methods}

\begin{table*}[t!]
\centering
\setlength{\tabcolsep}{2.8pt}
\renewcommand{\arraystretch}{1.15}

\resizebox{\textwidth}{!}{%
\begin{tabular}{l cc cc cc cc cc cc cc cc}
\toprule[1.5pt]
\rowcolor{headerblue}
\textbf{\textcolor{white}{Method}} &
\multicolumn{2}{c}{\textbf{\textcolor{white}{MedEE}}} &
\multicolumn{2}{c}{\textbf{\textcolor{white}{DisPA}}} &
\multicolumn{2}{c}{\textbf{\textcolor{white}{BillSC}}} &
\multicolumn{2}{c}{\textbf{\textcolor{white}{SocialTC}}} &
\multicolumn{2}{c}{\textbf{\textcolor{white}{ImageAR}}} &
\multicolumn{2}{c}{\textbf{\textcolor{white}{ImageCl}}} &
\multicolumn{2}{c}{\textbf{\textcolor{white}{ImageTR}}} &
\multicolumn{2}{c}{\textbf{\textcolor{white}{Average}}} \\
\rowcolor{headerblue}
\textbf{\textcolor{white}{}} &
\textcolor{white}{Acc} & \textcolor{white}{F1} &
\textcolor{white}{Acc} & \textcolor{white}{$\kappa$} &
\textcolor{white}{Acc} & \textcolor{white}{$\kappa$} &
\textcolor{white}{Acc} & \textcolor{white}{$\kappa$} &
\textcolor{white}{Acc} & \textcolor{white}{$\kappa$} &
\textcolor{white}{Acc} & \textcolor{white}{$\kappa$} &
\textcolor{white}{Acc} & \textcolor{white}{$\kappa$} &
\textcolor{white}{Acc} & \textcolor{white}{$\kappa$} \\
\midrule

Human 
& 82.76 & 82.09 
& 93.86 & 38.77 
& 94.66 & 83.50 
& 76.82 & 44.62 
& 94.06 & 58.59 
& 83.54 & 81.27 
& 96.86 & 56.37 
& 88.94 & 63.60 \\

\midrule
\multicolumn{17}{c}{\cellcolor{gray!30}\textbf{Single-Agent Method for Data Annotation}} \\
\midrule

Vanilla 
& 16.10 & 73.21 
& 91.80 & 36.28 
& 79.80 & 80.12 
& 54.20 & 41.63 
& 49.10 & 41.58 
& 40.20 & 76.19 
& 88.40 & 48.37 
& 59.94 & 56.77 \\

CoT~\cite{NEURIPS2022_9d560961} 
& 24.30 & 74.07 
& 92.40 & 38.59 
& 83.30 & 81.03 
& 57.80 & 42.92 
& 57.80 & 45.03 
& 45.10 & 77.20 
& 88.20 & 48.22 
& 64.13 & 58.15 \\

ICL~\cite{liu-etal-2022-makes} 
& 27.60 & 75.59 
& 91.60 & 40.13 
& 89.40 & 82.75 
& 63.30 & 44.26 
& 59.20 & 44.91 
& 54.60 & 78.37 
& 87.20 & 47.61 
& 67.56 & 59.09 \\

APE~\cite{zhou2023large} 
& 25.90 & 73.68 
& 91.50 & 40.96 
& 90.20 & 83.14 
& 52.90 & 41.45 
& 81.30 & 53.85 
& 60.20 & 79.12 
& 84.50 & 46.48 
& 69.50 & 59.81 \\

ORPO~\cite{yang2024large} 
& 28.10 & 73.60 
& 93.00 & 39.82 
& 84.20 & 81.53 
& 55.70 & 42.45 
& 81.30 & 54.59 
& 56.20 & 78.80 
& 83.60 & 45.31 
& 68.87 & 59.44 \\

ERM~\cite{yan-etal-2025-efficient} 
& 24.90 & 74.80 
& 92.80 & 40.27 
& 91.20 & 83.46 
& 50.00 & 40.12 
& 75.60 & 52.08 
& 55.80 & 78.57 
& 86.50 & 48.43 
& 68.11 & 59.68 \\

SPO~\cite{xiang2025self} 
& 28.20 & 74.44 
& 87.90 & 39.29 
& 88.00 & 82.46 
& 58.30 & 42.79 
& 79.30 & 53.56 
& 60.40 & 79.16 
& 86.20 & 47.76 
& 69.76 & 59.92 \\

\midrule
\multicolumn{17}{c}{\cellcolor{gray!30}\textbf{Multi-Agent Systems for Data Annotation}} \\
\midrule

Vanilla 
& 25.10 & 74.80 
& 92.80 & 40.25 
& 84.60 & 81.59 
& 60.40 & 43.48 
& 58.40 & 44.73 
& 48.00 & 77.73 
& 88.40 & 48.34 
& 65.39 & 58.70 \\

ICL~\cite{liu-etal-2022-makes} 
& 28.70 & 75.80 
& 92.50 & 40.83 
& 89.40 & 82.72 
& 63.30 & 44.44 
& 59.40 & 45.04 
& 56.10 & 78.68 
& 88.40 & 49.06 
& 68.26 & 59.51 \\

SPO~\cite{xiang2025self} 
& 27.70 & 74.78 
& 91.10 & 40.43 
& 90.50 & 83.18 
& 59.00 & 41.89 
& 80.00 & 54.02 
& 58.40 & 78.68 
& 85.00 & 46.56 
& 70.24 & 59.93 \\

\rowcolor{comprow}
ANT-OR 
& \textbf{35.90} & \textbf{76.10} 
& \textbf{93.50} & \textbf{41.05} 
& \textbf{92.60} & \textbf{83.82} 
& \textbf{65.00} & \textbf{44.50} 
& \textbf{86.40} & \textbf{56.61} 
& \textbf{60.80} & \textbf{79.22} 
& \textbf{89.00} & \textbf{50.15} 
& \textbf{74.74} & \textbf{61.64} \\

\bottomrule[1.5pt]
\end{tabular}}
\caption{
Performance comparison of different methods on data annotation tasks. 
We report Accuracy (agreement@$\ge$2) (\%) as the metric for annotation correctness, with Fleiss' Kappa ($\kappa$) measuring inter-annotator consistency. 
For Medical Entity Extraction, where Fleiss' $\kappa$ is not applicable due to span-level entity annotations, we use Entity-Level F1~\cite{medner2024} instead.
The best results are highlighted in \textbf{bold}.
}
\label{tab:main_results}
\end{table*}

\section{Experiments}
\subsection{Experimental Setup}
\paragraph{Models and Benchmarks.}
Considering model performance, cost, and inference efficiency, we conduct experiments using DeepSeek-V3.1-Terminus \cite{deepseekai2025deepseekv3} for text-modality tasks and Qwen3-235B-A22B-VL-Instruct \cite{bai2025qwen3vl} for vision-language tasks as the backbone models of MAS. Both models also serve as the optimizers themselves. As mentioned in Section~\ref{sec:benchmark}, we construct a benchmark covering multiple domains (medical, e-commerce, general) and multiple modalities (including text and vision-language tasks) to evaluate the performance of different optimization methods in data annotation scenarios.

\paragraph{Baselines.} We reproduce the relevant baselines in both single-agent and multi-agent system (MAS) scenarios. In single-agent scenarios, we compare with the Chain-of-Thought (CoT)~\cite{NEURIPS2022_9d560961} and In-Context Learning (ICL) via similarity-based example retrieval~\cite{liu-etal-2022-makes}, as well as other single-agent optimization algorithms, including APE~\cite{zhou2023large}, ORPO~\cite{yang2024large}, ERM~\cite{yan-etal-2025-efficient}, and SPO~\cite{xiang2025self}  (we reproduce the optimization logic of SPO in labeled scenarios). In multi-agent settings, we evaluate vanilla and ICL variants of the Annotator--Verifier architecture. We also adapt SPO, the strongest single-agent optimization baseline, to the same MAS architecture using the same batch schedule for a fair comparison. 

\paragraph{Implementation Details} 
We configure ANT-OR with 3 LLM-based annotators and one verifier, all at temperature=0.1. 
Each annotator is initialized with the same prompt template, while the verifier uses a separate adjudication prompt. 
Following the optimization procedure of SPO~\cite{xiang2025self}, we train on 200 samples per task with batch size 20, 
3 iterations per batch over 6 epochs, and select the best epoch based on training set performance. 
We evaluate on 1000 test samples; the remaining 300 samples are unused. 
For retrieval, we encode text and images with Qwen3-Embedding-8B~\cite{zhang2025qwen3embedding} and 
Qwen3-VL-Embedding-8B~\cite{li2026qwen3vlembedding}. Prompts are detailed in Appendix~\ref{sec:prompts}.

\paragraph{Metrics.}
We evaluate annotation performance with two complementary metrics:
\textbf{agreement@$\ge$2} measures whether an annotation is supported by at least two human annotators. 
For model predictions, we count a prediction as correct if it matches a label annotated by at least two of the five human annotators. 
For human performance, we use a leave-one-annotator setting: each human annotation is treated as the prediction and is counted as correct if at least one of the other four annotators gives the same label. 
\textbf{Fleiss' $\kappa$} measures inter-annotator consistency. 
We compute the human baseline from the five human annotators, and compute model $\kappa$ by treating the model as an additional annotator alongside the humans. 
For Medical Entity Extraction, where Fleiss' $\kappa$ is not applicable, we use \textbf{Entity-Level F1}~\cite{medner2024} instead.

\subsection{Main Results}

Table~\ref{tab:main_results} presents the comprehensive evaluation results comparing ANT-OR against single-agent and multi-agent baselines across all 7 datasets.

\paragraph{Consistent Superiority Across Tasks, Domains, and Modalities.}
ANT-OR achieves the best overall performance across all 7 datasets, outperforming the strongest baseline by 4.5 points in average agreement@$\ge$2 accuracy. The gains are consistent across medical, e-commerce, and general-domain tasks, as well as both text and vision-language modalities. These results show that ANT-OR can effectively handle nuanced decision boundaries in diverse guideline-governed annotation settings.

\paragraph{Narrowing the Gap to Human-Level Consistency.}
ANT-OR not only improves point-wise consensus coverage (agreement@$\ge$2), but also achieves higher distributional agreement with human judgments, as measured by Fleiss' $\kappa$. Notably, ANT-OR achieves $\kappa$ = 56.61 on ImageAR and $\kappa$ = 83.82 on BillSC, approaching the inter-annotator agreement among human annotators themselves (58.59 and 83.50 in Table~\ref{tab:benchmark_stats}). This validates that, compared to static agent systems, optimized multi-agent systems can narrow the gap to human-level consistency on complex annotation tasks.

\paragraph{Multi-Agent Stability and Effective Optimization.}
Multi-agent systems provide a more stable foundation for automated annotation: MAS consistently outperforms the basic single-agent CoT setup across all 7 datasets, with gains ranging from 0.2 to 2.9 points. Similarly, applying SPO to multi-agent systems improves average performance despite task-level variance.  ANT-OR fully realizes this potential through agent-level structured knowledge base, achieving gains over the vanilla MAS baseline on all seven datasets, establishing multi-agent optimization as a reliable strategy for improving automated annotation performance.

\begin{table}[t]
\centering
\scalebox{0.46}{
\begin{tabular}{l cc cc cc cc cc}
\toprule
\rowcolor{headerblue}
\textbf{\textcolor{white}{Method}} &
\multicolumn{2}{c}{\textbf{\textcolor{white}{MedEE}}} &
\multicolumn{2}{c}{\textbf{\textcolor{white}{DisPA}}} &
\multicolumn{2}{c}{\textbf{\textcolor{white}{BillSC}}} &
\multicolumn{2}{c}{\textbf{\textcolor{white}{SocialTC}}} & \multicolumn{2}{c}
{\textbf{\textcolor{white}{Avg.}}} \\
\rowcolor{headerblue}
\textbf{\textcolor{white}{}} &
\textcolor{white}{Acc} & \textcolor{white}{F1} &
\textcolor{white}{Acc} & \textcolor{white}{$\kappa$} &
\textcolor{white}{Acc} & \textcolor{white}{$\kappa$} &
\textcolor{white}{Acc} & \textcolor{white}{$\kappa$} &
\textcolor{white}{Acc} & \textcolor{white}{$\kappa$} \\
\midrule
\rowcolor{comprow}
ANT-OR (Full) & 35.90 & 76.10 & 93.50 & 41.05 & 92.60 & 83.82 & \textbf{65.00} & \textbf{44.50} & \textbf{71.75} & \textbf{61.37} \\
\midrule
\multicolumn{11}{c}{\cellcolor{gray!30}\textbf{I. Component Ablation}} \\
\midrule
w/o Dimension Mining & 34.90 & 75.93 & 91.20 & 40.34 & 87.60 & 82.01 & 58.50 & 42.59 & 68.05 & 60.22 \\
w/o Contrastive Triples & 32.80 & 75.70 & 89.50 & 40.35 & 91.90 & 83.69 & 63.70 & 43.18 & 69.47 & 60.74 \\
w/o Coupled Injection & 27.40 & 75.18 & 93.40 & 40.90 & \textbf{93.50} & \textbf{84.04} & 63.80 & 44.07 & 69.53 & 61.04 \\
\midrule
\multicolumn{11}{c}{\cellcolor{gray!30}\textbf{II. Effect of Injected Sample Number (Default=2)}} \\
\midrule
1 sample & 33.60 & 75.79 & \textbf{93.60} & \textbf{41.50} & 92.00 & 83.55 & 63.50 & 44.22 & 70.67 & 61.27 \\
3 samples & 35.40 & \textbf{76.36} & 93.10 & 40.95 & 92.20 & 83.66 & 64.20 & 44.17 & 71.22 & 61.29 \\
5 samples & \textbf{36.70} & 76.09 & 93.20 & 41.03 & 92.30 & 83.69 & 64.30 & 44.24 & 71.62 & 61.26 \\
\midrule
\multicolumn{11}{c}{\cellcolor{gray!30}\textbf{III. Effect of Retrieval Strategy}} \\
\midrule
Random & 34.80 & 75.87 & 92.80 & 40.85 & 91.80 & 83.59 & 63.80 & 44.38  & 70.80 & 61.17\\
Similarity only & 36.20 & 76.29 & 93.10 & 41.01 & 92.60 & 83.81 & 63.90 & 44.32 & 71.45 & 61.36 \\
\bottomrule
\end{tabular}
}
\caption{Ablation and sensitivity analysis of ANT-OR on four text datasets.}
\label{tab:ablation_all}
\end{table}

\begin{figure}[t]
  \includegraphics[width=\columnwidth]{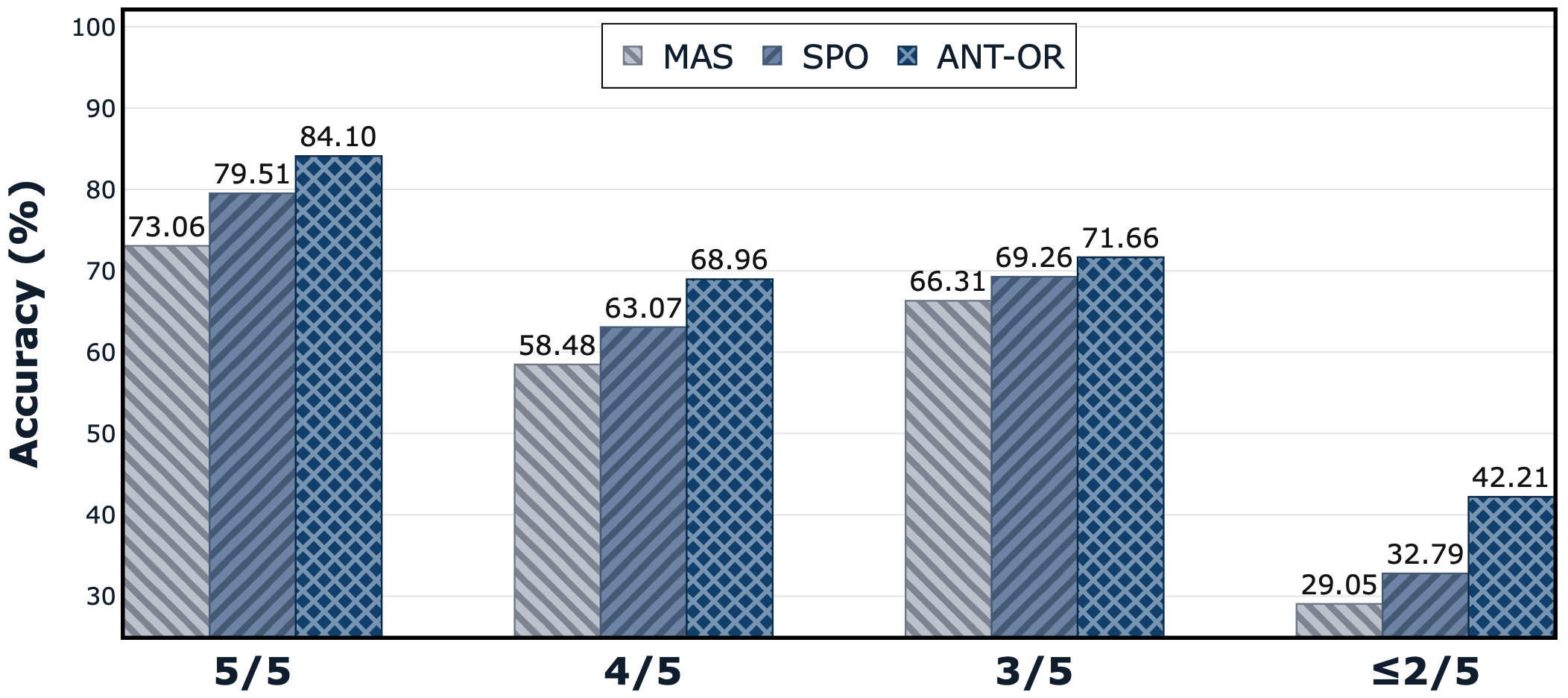}
  \caption{Average accuracy under different levels of human agreement.}
  \label{fig:agreement_analysis}
\end{figure}

\begin{figure}[t]
  \includegraphics[width=\columnwidth]{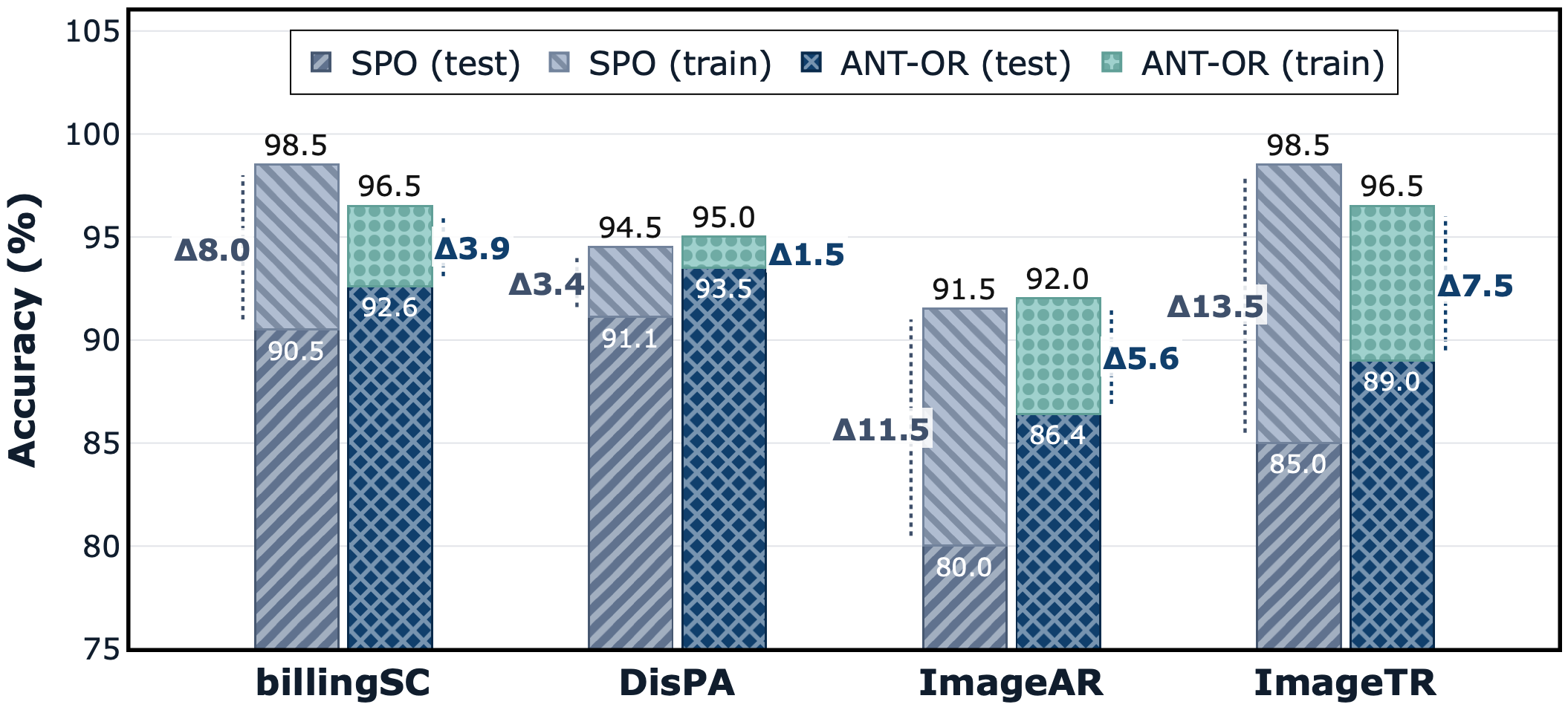}
  \caption{Generalization comparison of ANT-OR and SPO across datasets.}
  \label{fig:generalization}
\end{figure}

\subsection{Ablation Study}
\paragraph{Effect of Dimension Mining.}
Discriminative dimensions are the core knowledge units in ANT-OR, capturing recurring decision boundaries from annotation errors as reusable guidance. To evaluate their effect, we remove explicit dimension construction while keeping the rest of the pipeline unchanged. As shown in Table~\ref{tab:ablation_all}, performance drops consistently across all four datasets: average accuracy decreases from 71.75 to 68.05, and average $\kappa$ from 61.37 to 60.22. The larger decline on BillSC and SocialTC suggests that decision-level abstractions are important for modeling stable annotation criteria beyond instance-level corrections.

\paragraph{Effect of Contrastive Triples.}
We further examine the role of contrastive triples in the mining stage. In the full model, each error case is paired with semantically similar positive cases to expose fine-grained label distinctions. As shown in Table~\ref{tab:ablation_all}, removing this construction consistently hurts performance, lowering average accuracy from 71.75 to 69.47 and average $\kappa$ from 61.37 to 60.74. These results suggest that contrastive triples provide richer optimization signals than error cases alone.

\paragraph{Effect of Coupled Injection.}
We further examine coupled injection, which retrieves examples at inference time to contextualize discriminative dimensions. As shown in Table~\ref{tab:ablation_all}, it improves overall accuracy (69.53$\rightarrow$71.75) and $\kappa$ (61.04$\rightarrow$61.37), though gains vary across tasks. This reflects differences in boundary complexity: retrieved examples are more helpful when category distinctions are subtle, but contribute less to rule-based tasks and may introduce noise. These results suggest that discriminative dimensions drive the main improvements, while retrieved samples provide complementary inference-time support.

\paragraph{Effect of Injected Sample Number and Retrieval Strategy.}
Table~\ref{tab:ablation_all} shows that ANT-OR is relatively insensitive to the number of injected samples: varying from 1 to 5 yields only small performance changes. This suggests that injected examples provide contextual support, but their benefit saturates with few samples. For retrieval strategy, random selection performs worst, and our quality-aware similarity retrieval performs best overall, indicating that effective retrieval should consider both relevance and reliability.

\paragraph{Effect of positive samples.}

We use a 2:1 ratio of positive to negative samples in contrastive mining. Increasing the number of positive examples is intended to improve the accuracy of attribution.
Table~\ref{tab:ablation_positive} presents the ablation study on the number of positive samples used in contrastive mining. The ablation shows that moving from 1 to 2 positive samples yields clear gains on average agreement$\geq$2), while adding a third brings only marginal improvements. This confirms that two positives are sufficient to stabilize decision patterns without unnecessary cost, making the 2:1 ratio the optimal trade-off.

\begin{table}[t]
\centering
\scalebox{0.50}{
\begin{tabular}{lcc cc cc cc cc}
\toprule
\rowcolor{headerblue}
\textbf{\textcolor{white}{\# Pos}} & 
\multicolumn{2}{c}{\textbf{\textcolor{white}{BillSC}}} & 
\multicolumn{2}{c}{\textbf{\textcolor{white}{MedEE}}} & 
\multicolumn{2}{c}{\textbf{\textcolor{white}{Average}}} \\
\rowcolor{headerblue}
\textbf{\textcolor{white}{Samples}} & 
\textcolor{white}{Agree$\geq$2} & \textcolor{white}{Fleiss'$\kappa$} & 
\textcolor{white}{Agree$\geq$2} & \textcolor{white}{Fleiss'$\kappa$} & 
\textcolor{white}{Agree$\geq$2} & \textcolor{white}{Fleiss'$\kappa$} \\ 
\midrule
1 & 91.70 & 83.51 & 33.90 & 75.81 & 62.80 & 79.66 \\
2 & 92.60 & \textbf{83.82} & 35.90 & 76.10 & 64.25 & 79.96 \\
3 & \textbf{92.70} & 83.78 & \textbf{36.20} & \textbf{76.21} & \textbf{64.45} & \textbf{79.99} \\
\bottomrule
\end{tabular}
}
\caption{Ablation study on the number of positive samples in contrastive mining.}
\label{tab:ablation_positive}
\end{table}


\subsection{Analysis}

\paragraph{Analysis by Human Agreement Level.}
To assess performance under varying ambiguity, we group instances by maximum annotator agreement and report agreement@$\ge$2 averaged across datasets in Figure~\ref{fig:agreement_analysis}. Performance declines for all methods as agreement decreases. SPO offers limited gains on highly ambiguous cases ($\leq$2/5 agreement), suggesting that static optimization struggles with subtle decision boundaries. In contrast, ANT-OR improves substantially, outperforming MAS by nearly 8 points on this subset. This suggests that iterative contrastive optimization helps agents learn structured rules for guideline gray areas rather than merely fitting clear-cut cases.

\paragraph{Generalization Performance.} Figure~\ref{fig:generalization} compares training and test performance of ANT-OR and SPO. Although SPO achieves higher training accuracy (e.g., 98.5\% vs. 96.5\% on BillSC), ANT-OR performs better on the test sets in Figure~\ref{fig:generalization} by an average of 3.7 percentage points. This suggests that SPO tends to overfit the training data, while ANT-OR generalizes better to unseen instances.



\begin{table}[t]
\centering
\scalebox{0.50}{
\begin{tabular}{lcc cc cc cc cc}
\toprule
\rowcolor{headerblue}
\textbf{\textcolor{white}{Method}} &
\multicolumn{2}{c}{\textbf{\textcolor{white}{MedEE}}} &
\multicolumn{2}{c}{\textbf{\textcolor{white}{DisPA}}} &
\multicolumn{2}{c}{\textbf{\textcolor{white}{BillSC}}} &
\multicolumn{2}{c}{\textbf{\textcolor{white}{SocialTC}}} & \multicolumn{2}{c}
{\textbf{\textcolor{white}{Avg.}}} \\
\rowcolor{headerblue}
\textbf{\textcolor{white}{}} &
\textcolor{white}{Acc} & \textcolor{white}{Cost} &
\textcolor{white}{Acc} & \textcolor{white}{Cost} &
\textcolor{white}{Acc} & \textcolor{white}{Cost} &
\textcolor{white}{Acc} & \textcolor{white}{Cost} &
\textcolor{white}{Acc} & \textcolor{white}{Cost} \\ 
\midrule
SPO & 27.70 & 73.87 & 91.10 & 54.32 & 90.50 & 33.86 & 59.00 & 101.71 & 67.08 & 65.94\\
\rowcolor{comprow}
ANT-OR & \textbf{35.90} & 116.70 & \textbf{93.50} & 74.87 & \textbf{92.60} & 47.15 & \textbf{65.00} & 166.99 & \textbf{71.75} & 101.43\\ 
\bottomrule
\end{tabular}
}
\caption{Optimization Cost Analysis.}
\label{tab:optimization_cost}
\end{table}

\paragraph{Cost Analysis.} 
Table~\ref{tab:optimization_cost} compares optimization costs between SPO and ANT-OR. The optimization cost is calculated based on DeepSeek's official token pricing, converted to USD: \$0.26 per million input tokens (assuming 70\% of the input tokens hit the cache) and \$1.80 per million output tokens for DeepSeek-V3.1-Terminus. While ANT-OR incurs a higher average cost (\$101.43 vs. \$65.94), it achieves a substantial accuracy improvement of 4.7 points (71.75\% vs. 67.08\%). The additional cost stems from multi-agent contrastive mining, knowledge base construction and sample injection.

\section{Conclusion}
We present ANT-A, a multi-annotator benchmark and ANT-OR, a multi-agent framework, to advance automated data annotation toward human-level consistency. Our benchmark spans 7 datasets across multiple domains and modalities, providing 5 independent human annotations per sample alongside detailed guideline documents, thereby enabling rigorous evaluation of inter-annotator agreement. To address the limitations of static agent architectures, ANT-OR introduces iterative prompt optimization with a structured knowledge base and contrastive learning over annotation samples, mining discriminative dimensions from contrastive prediction patterns to transform collective annotation experience into reusable decision rules. Extensive experiments demonstrate that ANT-OR significantly narrows the gap to human-level consistency, consistently outperforming single-agent and multi-agent baselines across diverse tasks.

\section*{Limitations}
\paragraph{Broader Benchmark Coverage.}
Although our benchmark spans multiple domains and modalities, it still represents only a subset of real-world annotation settings. Future work can extend this benchmark to cover a wider range of tasks, such as long-document annotation, open-ended judgment, and more specialized expert workflows, in order to further evaluate automated annotation systems under broader and more realistic conditions.

\paragraph{More Efficient Optimization.}
ANT-OR improves annotation reliability through multi-agent collaboration and iterative optimization, but this design also introduces additional computational overhead. A natural direction for future work is to develop more efficient optimization and inference strategies that preserve the benefits of structured multi-agent refinement while reducing optimization and annotation cost.

\paragraph{Stronger and More General Backbone Models.}
The current framework is built on the capabilities of the underlying LLMs for retrieval, reasoning, and prompt refinement. As backbone models continue to improve, it will be important to study how ANT-OR scales to stronger models and how well it transfers across different model families, domains, and annotation settings.

\section*{Ethics Statement}
This work follows the ACL Ethics Policy. Our benchmark is constructed for research on reliable data annotation, and we take care to remove or anonymize personally identifiable or sensitive information before release. Human annotations are collected from professional annotators who are provided with task guidelines and clarification opportunities. Although ANT-OR improves LLM-based annotation reliability, it should not be used as a substitute for human experts in high-stakes domains such as medicine, finance, or legal decision-making. LLM-generated annotations may still contain errors or biases, so users should conduct task-specific privacy, bias, and reliability assessments before deployment.

\section*{Acknowledgments}
This work was supported in part by the Guangdong Basic and Applied Basic Research Foundation (Grant No. 2026B0101100004), and Shenzhen Science and Technology Program (Grant No. KJZD20231023094700001, KQTD2024072910215406).



\bibliography{custom}

\clearpage
\appendix
\label{sec:appendix}
\section{Multi-Agent Architecture Comparison}
\label{sec:architecture_comparison}

Table~\ref{tab:diff_structure_and_cost} compares different multi-agent system (MAS) architectures for data annotation. We analyze the failure modes of alternative designs:

\paragraph{Debate-based Architecture.}
Following \citet{du2024agentdebate}, we implement a multi-agent debate where agents iteratively discuss and refine their predictions. While debate improves reasoning on open-ended tasks, it performs poorly for guideline-governed annotation: (1) debates often diverge from the original task into tangential arguments, amplifying rather than correcting initial errors; (2) the lack of explicit role separation leads to groupthink, where agents converge on 
consensus rather than correct answers; (3) computational costs scale quadratically with debate rounds, making it economically impractical for large-scale annotation.

\paragraph{Planner-Worker Architecture.}
We implement a Planner-Worker design following standard task-decomposition patterns~\cite{wang2023plan}, where a planner agent divides inputs into sub-tasks and delegates to worker agents, then aggregates results. Contrary to expectations, this architecture underperforms single-agent baselines across all tasks (-1.1\% on MedEE, -3.7\% on DisPA, -0.4\% on BillSC). This failure may be attributed to coordination noise in guideline interpretation. Unlike simple task decomposition where sub-tasks are independent, annotation tasks require holistic understanding of interdependent category definitions and boundary rules. The planner's decomposition forces premature partitioning of decision logic: workers process truncated contexts that lack crucial disambiguating information from the full guidelines, while the aggregation step (simple concatenation or voting) lacks the capability to perform guideline-grounded adjudication. This creates a compounding error effect: the planner introduces interpretation errors at the decomposition stage, workers amplify these errors with incomplete context, and the aggregation fails to reconcile genuine disagreements. While costs are reduced compared to full debates (34-50\% lower), the accuracy degradation renders this architecture unsuitable for high-stakes annotation tasks.

\paragraph{Annotator-Verifier Architecture.}
Consistent with prior findings that annotator-verifier architectures enhance reliability~\cite{hegazy2025mafa,ahtisham2025orchestration}, ANT-OR uses parallel annotators to diversify perspectives while a dedicated verifier adjudicates conflicts through guideline-grounded reasoning. This design achieves 1-2\% accuracy improvement over baselines with only 3-4$\times$ cost increase (vs. 10-50$\times$ for alternatives), demonstrating superior accuracy-cost efficiency for annotation tasks.

\begin{table*}[t]
\centering
\begin{tabular}{@{}lcccccccc@{}}
\toprule
\textbf{Method} & \multicolumn{2}{c}{\textbf{MedEE}} & \multicolumn{2}{c}{\textbf{DisPA}} & \multicolumn{2}{c}{\textbf{BillSC}} & \multicolumn{2}{c}{\textbf{Average}}\\
& Acc & Cost & Acc & Cost & Acc & Cost & Acc & Cost\\
\midrule
CoT & 24.3 & 7.00 & 92.4 & 6.83 & 83.3 & 8.18 & 66.67 & 7.34\\
\midrule
Debate & 16.7 & 66.65 & 86.5 & 78.34 & 76.1 & 77.36 & 59.77 & 74.12\\
Planner-Worker & 23.2 & 37.07 & 88.3 & 34.11 & 82.9 & 33.31 & 64.80 & 34.83\\
Annotator-Verifier & \textbf{25.1} & 29.16 & \textbf{92.8} & 22.08 & \textbf{84.6} & 22.80 & \textbf{67.50} & 24.68\\
\bottomrule
\end{tabular}
\caption{Performance comparison on different MAS (w/o Prompt Optimization ).}
\label{tab:diff_structure_and_cost}
\end{table*}

\section{Case Study}
\label{sec:case_study}

In this appendix, we present two representative cases from Billing Scenario Classification Task, to illustrate (1) the realism of our benchmark and (2) the qualitative improvements brought by ANT-OR. To keep the case tables focused and readable, we present the annotation guideline for this task in the table below.

\paragraph{Case 1: A boundary instance with human disagreement.}
Table~\ref{tab:case1} shows an example where annotators disagree despite following the same annotation guideline. This case highlights an important property of real-world annotation: some instances lie near genuine decision boundaries, and their difficulty cannot be fully captured by a single gold label. By collecting five independent annotations per sample, our benchmark makes such ambiguity explicit and enables more realistic evaluation of annotation systems.

\begin{table*}[t]
\centering
\begin{tabular}{p{2.4cm} p{11.6cm}}
\toprule
\textbf{Task} & Billing Scenario Classification \\
\midrule
\textbf{Input} & 
\textbf{Merchant Name}: Leisure Tea House \newline
\textbf{Bill Title}: Leisure Tea House \newline
\textbf{Counterparty}: --- \\
\midrule
\textbf{Human Annotations} & [Food \& Dining, Culture \& Leisure, Food \& Dining, Food \& Dining, Culture \& Leisure] \\
\midrule
\textbf{Observation} & 
This case captures the inherent ambiguity of crossover business formats at the semantic boundary. According to the annotation rules, Food \& Dining covers ``beverages,'' while Culture \& Leisure encompasses ``leisure venue consumption, including cinemas, bars, chess and card rooms, etc.'' A tea house simultaneously operates as a beverage provider (Food \& Dining) and a leisure venue (Culture \& Leisure). Three annotators prioritized the beverage service aspect, whereas two annotators categorized it as a leisure venue analogous to a bar or chess room. This 3:2 split reflects the difficulty of assigning a single category to establishments whose core offering inherently spans multiple semantic boundaries. \\
\bottomrule
\end{tabular}
\caption{A representative boundary case with human disagreement in Billing Scenario Classification Task.}
\label{tab:case1}
\end{table*}

\paragraph{Case 2: Improvement after optimization.}
Table~\ref{tab:case2} compares the predictions of a vanilla multi-agent system (MAS) and ANT-OR on another difficult example from the same task. The baseline prediction is driven mainly by coarse lexical or topical cues, while ANT-OR gives the correct label by following the task guideline more precisely. This example qualitatively shows how iterative optimization improves annotation reliability.

\section{Benchmark Construction and Annotation Details}
\label{sec:benchmark_details}

This appendix provides comprehensive details on the construction, annotation process, and characteristics of the ANT-A benchmark, supplementing the descriptions in the main paper.

\subsection{Dataset Overview and Task Selection}
\label{sec:task_selection}

The construction of the seven tasks takes into account different modalities, domains, and scenario requirements. They are obtained by collecting suitable queries according to annotation guidelines used in real application scenarios and then re-annotating them. Most of the queries come from existing datasets or web sources, but they are re-annotated according to the guidelines of different tasks.

Our dataset consists of two parts: annotation guidelines and queries. We first created the annotation guidelines, which were adapted from actual annotation requirements used in real-world scenarios, rather than being derived from existing datasets or manually fabricated from scratch. For example, the billing scenario classification task is used for bill-category identification in payment scenarios, and image-text relevance can be used in image search scenarios. We then collected corresponding queries according to the tasks described in the guidelines. After organizing the guidelines and queries, we manually annotated the queries according to the guidelines.

Table~\ref{tab:dataset_overview} presents detailed information about the seven tasks.

\begin{table*}[t]
\centering
\small
\setlength{\tabcolsep}{3.5pt}
\begin{tabular}{@{}lllp{3.5cm}p{5.5cm}@{}}
\toprule
\textbf{Dataset} & \textbf{Domain} & \textbf{Modality} & \textbf{Task Usage} & \textbf{Query Source} \\
\midrule
BillSC & E-commerce & Text & Identifying bill types in payment scenarios & Anonymized billing records \\
MedEE & Medical & Text & Medical Content Analysis and Structuring & Huatuo-26M~\cite{li2023huatuo} \\
DisPA & Medical & Text & Privacy Protection and Content Review & QABasedOnMedica\-KnowledgeGraph (GitHub) \\
SocialTC & General & Text & Knowledge Classification and Content Layering & Zhihu Website \\
ImageTR & General & Vision-Language & Search images by text & Paraphrased from MUGE~\cite{lin2021m6} via LLM; images via Bing Image Search API \\
ImageCI & E-commerce & Vision-Language & Sorting and managing poster materials & CGL-Dataset~\cite{li2023relation} \\
ImageAR & General & Vision-Language & Score Image Identifiability & Products-10K~\cite{bai2020products10k} \\
\bottomrule
\end{tabular}
\caption{Overview of the seven tasks in the ANT-A benchmark.}
\label{tab:dataset_overview}
\end{table*}

\subsection{Annotation Guidelines and Query Characteristics}
\label{sec:guideline_details}

The annotation guidelines are adapted from actual annotation requirements used in real production annotation scenarios (see the Task Usage column in Table~\ref{tab:dataset_overview} for their actual applications), rather than being manually fabricated from scratch. For example, the billing scenario classification task is used for bill-category identification in real payment scenarios, and the image-text relevance task is used in image search scenarios. The queries in different datasets are deduplicated and anonymized. We use a locally deployed DeepSeek-V3.1-Terminus to evaluate and anonymize sensitive information, and further conduct manual spot checks and a pre-annotation stage to ensure that the queries are suitable for annotation under the tasks specified by the guidelines.

It should be emphasized that the complexity of real annotation tasks does not necessarily equate to extremely long guidelines. Many real-world production annotation tasks are based on operational rules centered around specific business goals, such as bill-type identification, image-text relevance judgment, or image recognizability assessment. These guidelines may not be hundreds of pages long, but they require annotators to handle a large number of boundary cases in real samples. Our guidelines are derived from such real annotation requirements. They specify the task objectives, category boundaries, and judgment criteria, but the real difficulty lies in the fact that guidelines can hardly exhaustively cover all possible sample cases: sample information may be incomplete, categories may overlap, and the text-image pair may be supported by only partial evidence. As a result, multiple annotators may provide different yet reasonable judgments. We retain five independent annotation results precisely to characterize this real-world uncertainty. Therefore, annotator disagreement itself is part of the value of the benchmark.

\begin{table*}[t]
\centering
\small
\begin{tabular}{@{}lccccc@{}}
\toprule
\textbf{Dataset} & \textbf{\shortstack{Avg Guideline\\Length (chars)}} & \textbf{\shortstack{Avg Query\\Length (chars)}} & \textbf{4--5 Agree (\%)} & \textbf{2--3 Agree (\%)} & \textbf{No Agree (\%)} \\
\midrule
BillSC & 9,236 & 160 & 86.1 & 13.9 & 0.0 \\
MedEE & 4,624 & 261 & 61.1 & 33.9 & 5.0 \\
DisPA & 4,499 & 22 & 80.1 & 19.9 & 0.0 \\
SocialTC & 27,529 & 2,700 & 44.8 & 53.3 & 1.9 \\
ImageTR & 3,907 & 32 & 89.9 & 10.1 & 0.0 \\
ImageCI & 7,668 & N/A & 82.5 & 17.5 & 0.0 \\
ImageAR & 12,487 & N/A & 62.3 & 36.6 & 1.1 \\
\bottomrule
\end{tabular}
\caption{Statistics of annotation guidelines, query lengths, and inter-annotator agreement distributions. We define 4--5 agreement as cases where at least four annotators selected the same label, 2--3 agreement as cases where the majority label was selected by two or three annotators, and no agreement as cases without a dominant label.}
\label{tab:guideline_stats}
\end{table*}

Table~\ref{tab:guideline_stats} presents detailed statistics about the guidelines, queries, and annotation agreement distributions. The lower agreement in SocialTC and ImageAR is expected, as these tasks involve hierarchical semantic classification, subjective content layering on open-domain social text, and visual identifiability assessment, where multiple labels can be defensible due to inherently ambiguous semantic boundaries, a property that single-label benchmarks collapse but our multi-annotator design explicitly preserves.

\subsection{Annotation Process}
\label{sec:annotation_process}

Our three-stage annotation process is designed to differentiate meaningful disagreement from ambiguity caused by unclear guidelines.

\paragraph{Phase 1--2: Pilot Annotation and Guideline Refinement.}
We use approximately 50 pilot samples to expose ambiguities and missing rules, which are resolved through Q\&A and guideline revision before formal annotation begins.

\paragraph{Phase 3: Formal Annotation.}
After guideline refinement, formal annotation is conducted by five independent annotators per sample.

This process is further supported by the annotation statistics shown in Table~\ref{tab:guideline_stats}: the proportion of instances where all 5 annotators give completely different labels is about 1\% on average. If disagreements were primarily caused by unclear guidelines, we would expect a much higher rate of random divergence. The fact that full disagreement is rare indicates that most samples reach a consensus under the refined guidelines, and the retained disagreements are concentrated in specific boundary cases that resist unique adjudication. Thus, the disagreements in our experiments reflect differences that persisted after guideline refinement, not systematic ambiguity.

\subsection{Annotator Demographics and Quality Control}
\label{sec:annotator_demographics}

All annotators are recruited from a professional annotation company. Table~\ref{tab:annotator_info} presents the background information of the human annotators.

\begin{table*}[t]
\centering
\small
\begin{tabular}{@{}ccccc@{}}
\toprule
\textbf{ID} & \textbf{Age} & \textbf{Degree} & \textbf{Expertise} & \textbf{Years of Experience} \\
\midrule
1 & 37 & Bachelor & Law & 6 \\
2 & 33 & Bachelor & Computer Application Technology & 2.5 \\
3 & 29 & Bachelor & Business Administration & 2.5 \\
4 & 25 & Bachelor & Electronic Information Engineering Technology & 2 \\
5 & 27 & Bachelor & Computer Science and Technology & 2 \\
6 & 26 & Bachelor & E-Commerce & 2 \\
7 & 28 & Bachelor & Public Administration / Administrative Management & 2 \\
8 & 30 & Bachelor & Human Resource Management & 2 \\
9 & 28 & Bachelor & Electronic Information Engineering & 2 \\
10 & 28 & Bachelor & Land Resource Management & 2 \\
\bottomrule
\end{tabular}
\caption{Demographics of the human annotators.}
\label{tab:annotator_info}
\end{table*}

All annotators hold a bachelor's degree and have at least two years of annotation experience. Their competence is verified through the pilot annotation phase before formal labeling. We note that the two medical-related tasks (DisPA and MedEE) do not require professional medical expertise; their annotation objectives are grounded in explicit textual patterns and predefined schemas, which general annotators can accurately perform after being trained with the detailed guidelines.

\subsection{Qualitative Analysis of Prompt Optimization}
\label{sec:qualitative_prompt}

To complement the quantitative results, we provide a qualitative analysis of how the prompts evolve through the optimization process. Below is a representative example from the Billing Scenario Classification task, illustrating how the annotation prompt improves across optimization epochs.

\paragraph{Example: Merchant Name Specificity Threshold in Billing Scenario Classification.} \quad \\

\textbf{Epoch 0 (Simple condition):}
\begin{quote}
``When merchant name contains explicit industry keywords, classify by industry attribute.''
\end{quote}

\textbf{Epoch 2 (Added anti-examples):}
\begin{quote}
``When merchant name is too generic (e.g., only contains `company', `industry', `group' without industry indication) AND transaction description is ambiguous $\rightarrow$ classify as `Other'.''
\end{quote}

\textbf{Epoch 5 (Balanced two-tier rule):}
\begin{quote}
``When merchant name contains explicit industry keywords $\rightarrow$ prioritize industry over transaction nature; when overly generic + ambiguous content $\rightarrow$ `Other'.''
\end{quote}

\textbf{Why it matters:} This evolution prevents both false positives (over-eager industry classification) and false negatives (missing valid industry signals), allowing the model to better align with human understanding.

\section{Prompts}
\label{sec:prompts}
We present all the prompts used in our paper in the following tables, including:
\begin{itemize}
    \item \textbf{Prompts for Basic Annotation (Single Annotator):} The prompt is used to perform standard annotation when no contrastive samples or dimensions are available. The system provides the base annotation rules and the query to be annotated, then the LLM generates the annotation result with step-by-step reasoning.

    \item \textbf{Prompts for Dimension-Aware Annotation with Contrastive Samples:} The prompt is used to perform annotation when discriminative dimensions and contrastive samples are retrieved from the example pool. The system provides the base rules, relevant dimensions (rule intent, decision boundary, applicable boundary), and reference samples (both correct and incorrect demonstrations) to guide the LLM in making more accurate annotations by learning from historical error patterns.

    \item \textbf{Prompts for Verification Agent:} The prompt is used to select the best annotation result from multiple annotators by comparing their reasoning logic. When annotators disagree, the LLM evaluates each annotator's reasoning for logical consistency, coherence, and errors, then selects the most sound reasoning. This serves as a meta-judge to resolve disagreements among the annotator ensemble.

    \item \textbf{Prompts for Contrastive Dimension Mining:} The prompt is used to analyze contrastive triples (two correctly judged samples and one incorrectly judged sample that are semantically similar) to uncover the fundamental intent behind rule formulation. The LLM identifies surface differences, decision boundaries, rule intent, meta-rules, and applicable boundaries, producing structured dimension definitions that capture the core logic distinguishing correct from incorrect judgments.

    \item \textbf{Prompts for Dimension Field Refinement:} The prompt is used to integrate newly analyzed contrastive cases with existing discriminative dimensions. The LLM determines the relationship between new cases and historical rules (support, supplement, correction, or extension), extracts generalizable rule intent, resolves conflicts, and tests counterexamples to produce refined rule\_intent, decision\_boundary, and applicable\_boundary definitions.

    \item \textbf{Prompts for Verifier Prompt Optimization:} The prompt is used to optimize the verification agent's selection criteria based on error analysis. The system provides current verification errors, learned discriminative dimensions, and contrastive samples as auxiliary reference. The LLM analyzes root causes of verification failures and produces an improved verification prompt that can more accurately select the best annotator result by checking reasoning logic.

    \item \textbf{Prompts for Annotator Prompt Optimization:} The prompt is used to optimize the annotation instructions based on error analysis. The system provides the existing annotation prompt, current error samples (core optimization basis), learned discriminative dimensions, and contrastive samples as auxiliary reference. The LLM analyzes root causes of annotation errors and produces an improved annotation prompt that directly addresses the identified problems while following optimization principles.
\end{itemize}

\begin{table*}[t]
\centering
\begin{tabular}{p{2.4cm} p{11.6cm}}
\toprule
\textbf{Task} & Billing Scenario Classification \\
\midrule
\textbf{Input} & 
\textbf{Merchant Name:} Convenience Retail Store \newline
\textbf{Bill Title:} Personal Transfer Receipt \newline
\textbf{Counterparty:} Individual Customer \\
\midrule
\textbf{Human Annotations} & [Daily Necessities, Daily Necessities, Daily Necessities, Daily Necessities, Daily Necessities] \\
\midrule
\textbf{MAS} & \textbf{Result:} Transfer \& Red Envelope \newline
\textbf{Reasoning:} Focused on the surface-level keyword ``transfer'' in the bill title and treated the transaction as a peer-to-peer fund transfer between individuals, completely overlooking the merchant's industry attribute implied by ``Retail Store.'' \\
\midrule
\textbf{ANT-OR} & \textbf{Result:} Daily Necessities \newline
\textbf{Reasoning:} Applied the dimension-based rule that merchant names containing explicit industry keywords (e.g., ``retail store'') should take priority over payment-behavior descriptions in the title. Referenced analogous cases where merchants with clear industry labels were correctly classified by business type despite transfer/receipt descriptors. \\
\midrule
\textbf{Observation} & 
This case demonstrates a classic failure mode of literal keyword matching. MAS is misled by the word ``transfer'' in the bill title and ignores the structural priority between merchant industry and transaction behavior. In contrast, ANT-OR's multi-dimensional reasoning framework correctly elevates the merchant name's explicit industry signal (``retail store'' $\rightarrow$ \textit{Daily Necessities}) over the payment-operation description. The unanimous human consensus (5/5) on \textit{Daily Necessities} confirms that ANT-OR's reasoning aligns with human judgment while MAS falls into the trap of superficial pattern matching. \\
\bottomrule
\end{tabular}
\caption{Qualitative comparison between MAS and ANT-OR on Billing Scenario Classification Task.}
\label{tab:case2}
\end{table*}

\begin{table*}
\begin{tcolorbox}[notitle, colback=gray!10,
colframe=black,
title={Billing Scenario Classification Guidelines (Part 1)},]
\texttt{\\
\# Bill Scenario Classification Task - Annotation Rules Document\\
\\
\#\# I. Rules and Examples\\
\\
\#\#\# 1.1 Annotation Basics\\
\\
Categorization should be determined based on a comprehensive assessment of the transaction content. The transaction content must clearly correspond to the definition of a specific category; otherwise, it should be classified as ``Others.''\\
\\
\#\#\# 1.2 Data Description\\
\\
Transaction Content: Describes the specifics of the current transaction, indicating its exact nature.\\
\\
\#\#\# 1.3 Annotation Instructions\\
\\
Category Names and Definitions\\
\\
| Level-1 Category Name | Level-1 Category Semantics|\\
| ------------------------- | ----------------------------------- |\\
| Food \& Dining| Encompasses meals, fast food, beverages, fresh food, snacks, and tobacco-related consumption and services.|\\
| Apparel \& Accessories | Retail consumption of men's and women's clothing, shoes, bags, and jewelry, covering specialty stores, e-commerce platforms, and accessory brands. |\\
| Daily Necessities | Retail services for daily necessities, including department stores, convenience stores, flowers and plants, and community group-buying consumption.|\\
| Home Furnishings \& Renovation | Covers furniture, building materials, and renovation services; provides home textiles, lighting, and other household products to optimize living spaces. |\\
| Electronics \& Appliances| Covers major appliances, small appliances, and 3C digital product purchases and related consumption.|\\
| Sports \& Outdoors | Covers gym memberships, yoga and dance, sports venues, and outdoor sports equipment-related consumption.|\\
| Beauty \& Hair Care| Covers beauty and hair care-related consumption, including beauty and personal care, skincare, facial masks, nail art, as well as haircuts, perms, and coloring services and products. |\\
| Mother \& Baby | Consumption of infant and child products, children's clothing, toys, and related maternal and infant goods and services. |\\
}
\end{tcolorbox}
\end{table*}

\begin{table*}
\begin{tcolorbox}[notitle, colback=gray!10,
colframe=black,
title={Billing Scenario Classification Guidelines (Part 2)},]
\texttt{ \\
| Pets | Covers pet-related service consumption, including pet supplies, pet toys, pet hospitals, pet clothing, and grooming services.|\\
| Transportation| Covers transportation and travel consumption, including intercity transport such as air tickets, ferry tickets, train tickets, and bus tickets, as well as urban travel such as subway, bus, and toll fees.|\\
| Vehicle Maintenance | Fuel and charging, maintenance and repair, and parking services; covers auto repair shops, gas stations, charging stations, and sales of car/bicycle accessories.|\\
| Housing \& Property| Rent, property fees, and agency services; covers property management, real estate leasing, and agency services.| \\
| Hotels \& Travel| Covers hotel accommodation and travel-related consumption, including scenic spot tickets, hotels and homestays, travel packages, tour services, and annual passes.| \\
| Culture \& Leisure| Covers film and entertainment, performances and events, and leisure venue consumption, including cinemas, bars, chess and card rooms, etc., for both offline venues and online platforms.| \\
| Education \& Training | Covers education and training-related consumption, including tuition payment, online course purchases, public and private educational institutions, and training services. |\\
| Medical \& Health | Covers hospitals and clinics, physical examination and health care, medical beauty, medical equipment, and nutritional supplements-related services and consumption. | \\
| Lifestyle Services | Addresses transactional life needs, such as rentals, logistics, housekeeping and repairs, same-city delivery, legal and wedding services, and other specialized service fees. | \\
| Public Services | Fees for administrative, judicial, social security, housing provident fund, and other public affairs management provided by government agencies, as well as non-profit organization services. | \\
| Business Services | Fees arising from corporate operational support such as finance, transportation, marketing, and tax consulting, as well as e-commerce transaction guarantee services. | \\
| Charity \& Donations                    | Transactions for charitable donations, such as foundations (e.g., One Foundation), charity platforms (e.g., Shuidichou), or temples (e.g., incense donations), or charitable fundraising (e.g., rural primary schools). | \\
| Mutual Aid \& Protection | Payments for mutual aid protection plans, such as online mutual aid (e.g., Xianghubao sharing) or community insurance (e.g., neighborhood medical fund) consumption. | \\
| Investment \& Wealth Management | Covers financial product purchases, investment services, and financial institution-related transactions and expense payments.|\\
| Insurance | Covers various insurance products and services, including policy purchases, claims, compensation, and related protection expenses. | \\  
}
\end{tcolorbox}
\end{table*}

\begin{table*}
\begin{tcolorbox}[notitle, colback=gray!10,
colframe=black,
title={Billing Scenario Classification Guidelines (Part 3)},]
\texttt{ \\
| Credit \& Rental | Covers shared rentals, lending platforms, and financial leasing-related services and expense payments. | \\
| Recharge \& Utilities | Covers utilities (water, electricity, gas), communication, and network recharge services, including public utility payments, phone credit and data top-ups, and telecommunications operator services. | \\    
| Income | Various types of interest income. | \\
| Transfer \& Red Envelope | Content related to transfers and red envelopes. | \\
| Payment on Behalf of Relatives/Friends | Content related to payments made on behalf of relatives or friends. | \\
| Account Deposit \& Withdrawal | Content related to account deposits and withdrawals. | \\
| Refund | Transactions explicitly containing the word ``refund'' in the title, or other refund bills. | \\
| Others | Content that does not belong to any of the above level-1 categories. | \\
}
\end{tcolorbox}
\end{table*}


\begin{table*}
\begin{tcolorbox}[notitle, colback=gray!10, colframe=black, title={Prompts for Basic Annotation (Single Annotator)},]
\texttt{\\
System: You are a helpful assistant.\\
\\
User:\\
\{prompts/guidelines\}\\
\\
\#\#\# Query to be Annotated:\\
\{query\}\\
\\
\#\#\# Please output in the following format:\\
Reasoning Process: [Step-by-step reasoning]\\
Annotation Result: \{output\_format\}\\
}
\end{tcolorbox}
\end{table*}

\begin{table*}
\begin{tcolorbox}[notitle, colback=gray!10, colframe=black, title={Prompts for Dimension-Aware Annotation with Contrastive Samples (Part 1)},]
\texttt{\\
System: You are a helpful assistant.\\
\\
User:\\
\{prompts/guidelines\}\\
\\
\#\#\# Key Discriminative Dimensions (Summarized from Historical Errors)\\
The following dimensions are associated with the provided reference samples. Please annotate by combining dimension understanding with sample demonstrations:\\
\\{}
[Dimension 1: \{dim\_name\}]\\
- Rule intent: \{rule\_intent\}\\
- Decision boundary: \{decision\_boundary\}\\
- Applicable scenario: \{applicable\_boundary\}\\
\\{}
[Dimension 2: ...]\\
\\
\#\#\# Reference Samples (Associated with the Above Dimensions)\\
\\{}
[Reference Sample 1 (Correct Demonstration)] (Related dimensions: \{dim\_names\}) [Quality Tag] (similarity: \{sim\_score\})\\
Query: \{query\}\\
Annotation result: \{annotation\}\\
Correct answer: \{ground\_truth\}\\
\\{}
[Reference Sample 2 (Incorrect Demonstration)] (Related dimensions: \{dim\_names\}) [Quality Tag] (similarity: \{sim\_score\})\\
Query: \{query\}\\
Incorrect annotation: \{annotation\}\\
Correct answer: \{ground\_truth\}\\
Error reason: \{error\_reason\}\\
\\{}
[Contrastive Analysis Guidance]\\
Before making your judgment, carefully compare:\\
1. Core similarities between the current query and the correct sample(s)\\
2. Key differences between the current query and the incorrect sample(s)\\
3. How to avoid the mistakes made in the incorrect samples\\
4. Look for consistent patterns across multiple samples to improve judgment confidence\\
\\
\#\#\# Usage Guidance\\
1. First understand the rule intent and decision boundary of key discriminative dimensions\\
2. Refer to correct samples to understand how to judge, refer to incorrect samples to learn common pitfalls\\
3. Combine with the rule document to make an independent judgment on the current Query\\
}
\end{tcolorbox}
\end{table*}

\begin{table*}
\begin{tcolorbox}[notitle, colback=gray!10, colframe=black, title={Prompts for Dimension-Aware Annotation with Contrastive Samples (Part 2)},]
\texttt{
\\
\#\#\# Query to Annotate:\\
\{query\}\\
\\
\#\#\# Please output in the following format:\\
Reasoning process: [Step-by-step analysis of Query features, relevant dimensions, and comparison with reference samples]\\
Annotation result: \{output\_format\}\\
}
\end{tcolorbox}
\end{table*}

\begin{table*}
\begin{tcolorbox}[notitle, colback=gray!10, colframe=black, title={Prompts for Final Judgment Aggregation (Part 1)},]
\texttt{\\
System: You are a helpful assistant.\\
\\
User:\\
You are a data annotation expert. Your task is to carefully analyze the query to be annotated and multiple candidate annotation results, and provide the final annotation result for the query.\\
\\
\#\#\# Original Rule Document:\\
\{doc\_content\}\\
\\
\#\#\# Query to be Annotated:\\
\{query\}\\
\\
\#\#\# Candidate Annotation Results:\\
Annotation result 1:\\
\{annotation\_result\}\\
\\
Annotation result 2:\\
\{annotation\_result\}\\
\\
Annotation result 3:\\
\{annotation\_result\}\\
\\
Please note that the rule document serves only as a simple reference. Please carefully compare multiple candidate annotations and provide the final annotation result based on these candidate results.\\
\\
\#\#\# Judgment Suggestions:\\
\#\#\# Key Discriminative Dimensions (Summarized from Historical Errors)\\{}
[Dimension 1: \{dim\_name\}]\\
- Rule intent: \{rule\_intent\}\\
- Decision boundary: \{decision\_boundary\}\\
- Applicable scenario: \{applicable\_boundary\}\\
}
\end{tcolorbox}
\end{table*}

\begin{table*}
\begin{tcolorbox}[notitle, colback=gray!10, colframe=black, title={Prompts for Final Judgment Aggregation (Part 2)},]
\texttt{\\
\#\#\# Similar Sample Reference\\{}
[Reference Sample 1 (Correct Demonstration)]\\
Query: \{query\}\\
Annotation result: \{annotation\}\\
Correct answer: \{ground\_truth\}\\
\\{}
[Reference Sample 2 (Incorrect Demonstration)]\\
Query: \{query\}\\
Incorrect annotation: \{annotation\}\\
Correct answer: \{ground\_truth\}\\
Error reason: \{error\_reason\}\\
\\
\#\#\# Please output in the following format:\\
Analysis Process: [Your analysis of the query to be annotated and candidate annotation results]\\
Annotation Result: \{output\_format\}\\
}
\end{tcolorbox}
\end{table*}

\begin{table*}
\begin{tcolorbox}[notitle, colback=gray!10, colframe=black, title={Prompts for Verification Agent},]
\texttt{\\
System: You are a helpful assistant.\\
\\
User:\\
Check the reasoning logic of multiple annotators and select the best result.\\
\\
Focus on:\\
1. Compare the reasoning logic of each annotator\\
2. Check for logical consistency and coherence\\
3. Identify any reasoning errors or inconsistencies\\
4. Select the annotator with the most sound reasoning\\
\\
Note: When all three annotators disagree (three-way tie), you must select the best one based on reasoning quality.\\
\\
Query: \{query\_str\}\\
\\{}
[Key Discriminative Dimensions]\\{}
[Dimension 1: \{dim\_name\}]\\
- Rule intent: \{rule\_intent\}\\
- Decision boundary: \{decision\_boundary\}\\
- Applicable scenario: \{applicable\_boundary\}\\
\\
Annotations:\\
=== annotator1 (label: \{label\}) ===\\
\{annotation\_result\}\\
\\
=== annotator2 (label: \{label\}) ===\\
\{annotation\_result\}\\
\\
=== annotator3 (label: \{label\}) ===\\
\{annotation\_result\}\\
\\
Vote Info: Majority vote: \{winning\_label\} (\{winning\_count\}/\{total\_votes\})\\
OR Vote: Three-way tie (1:1:1), no majority\\
\\{}
[Similar Sample Reference]\\{}
[Reference samples with correct/incorrect demonstrations]\\
\\
Output format:\\
```json\\
\{\\
  "reason": ["logic error description for each annotator"],\\
  "result": ["annotator1/annotator2/annotator3"]\\
\}
```
}
\end{tcolorbox}
\end{table*}

\begin{table*}
\begin{tcolorbox}[notitle, colback=gray!10, colframe=black, title={Prompts for Contrastive Dimension Mining (Part 1)},]
\texttt{\\
System: You are a helpful assistant.\\
\\
User:\\
Analyze this group of cases that are 'similar but judged differently' to deeply uncover the fundamental intent behind the rule formulation.\\
\\
These samples are very similar in semantic input, but some are correctly judged while others are incorrectly judged.\\
\\
\#\#\# Original Rule Document:\\
\{doc\_content\}\\
\\
\#\#\# Historical Discriminative Dimensions:\\
Before inducing new dimensions, please check the following existing discriminative dimensions. If you find that the discriminative dimension reflected in the current case is substantially the same as a historical dimension (i.e., describing the same type of meta-rule), please directly use that historical dimension's meta-rule as the name instead of creating a new name.\\
\\
\{existing\_dims\_text\}\\
\\
\#\#\# Case group to analyze:\\{}
[Contrastive Group]\\
Sample A (Correctly judged):\\
  Query: \{query\}\\
  Model output: \{annotation\}\\
  Correct answer: \{ground\_truth\}\\
\\
Sample B (Correctly judged):\\
  Query: \{query\}\\
  Model output: \{annotation\}\\
  Correct answer: \{ground\_truth\}\\
\\
Sample C (Incorrectly judged):\\
  Query: \{query\}\\
  Model output: \{annotation\}\\
  Correct answer: \{ground\_truth\}\\
\\
Please deeply analyze this contrastive case group to uncover the fundamental intent behind rule formulation:\\
}
\end{tcolorbox}
\end{table*}

\begin{table*}
\begin{tcolorbox}[notitle, colback=gray!10, colframe=black, title={Prompts for Contrastive Dimension Mining (Part 2)},]
\texttt{
\\{}
1. [Surface Differences] What are the literal differences between samples? Especially between correct and incorrect samples\\{}
2. [Decision Boundary] What factors cause samples to be judged as correct or incorrect? (How should similar samples be annotated?)\\{}
3. [Rule Intent] What is the core reason why rule makers require this annotation? (i.e., what is the general core annotation logic?)\\{}
4. [Meta-Rule] If this distinguishing principle is abstracted into a general rule, how should it be expressed?\\{}
5. [Applicable Boundary] Under what circumstances does this rule apply? When might it not apply?\\
\\
Please note: Please analyze carefully based on the samples themselves. Since some samples may have deviations, please avoid using overly absolute judgment logic.\\
\\
Please output your complete reasoning process first, then output the final JSON result:\\
\\
Reasoning process: [Your reasoning process]\\
\\
Output JSON format:\\
```json\\
\{\\
  "surface\_difference": "Description of surface differences",\\
  "decision\_boundary": "Decision boundary (How should similar samples be annotated?)",\\
  "rule\_intent": "Core logic of rule formulation (i.e., general task annotation logic)",\\
  "meta\_rule": "Reusable meta-rule expression (used as dimension name, prioritize using historical dimension names)",\\
  "applicable\_boundary": "Applicable boundary conditions (what types of samples it applies to)"\\
\}\\
```\\
\\
Note:\\
- meta\_rule will be used as the unique identifier name for the discriminative dimension; if it is substantially the same as a historical dimension, please use the historical dimension name\\
}
\end{tcolorbox}
\end{table*}

\begin{table*}
\begin{tcolorbox}[notitle, colback=gray!10, colframe=black, title={Prompts for Dimension Field Refinement (Part 1)},]
\texttt{ \\
System: You are a helpful assistant.\\
\\
User:\\
You are an annotation rule expert. Based on the main rule document, integrate historical rules with new cases to extract generalizable rule intent (core annotation logic).\\
\\
\#\#\# Definitions\\
\\
1. [Rule Intent] What is the core reason why rule makers require this annotation? (i.e., the general core annotation logic)\\
2. [Decision Boundary] What factors cause one case to be judged as X and another as Y?\\
3. [Applicable Boundary] Under what circumstances does this rule apply? When might it not apply?\\
\\
\#\#\# Input Information\\
\\{}
[Original Rule Document]\\
\{doc\_content\}\\
\\{}
[Historical Discriminative Dimensions]\\
- Rule Intent: \{existing\_intent\}\\
- Decision Boundary: \{existing\_boundary\}\\
- Applicable Boundary: \{existing\_applicable\}\\
\\{}
[New Case Analysis]\\{}
[Contrastive Group]\\
Sample A (Correctly judged):\\
  Query: \{query\}\\
  Model output: \{annotation\}\\
  Correct answer: \{ground\_truth\}\\
\\
Sample B (Correctly judged):\\
  Query: \{query\}\\
  Model output: \{annotation\}\\
  Correct answer: \{ground\_truth\}\\
\\
Sample C (Incorrectly judged):\\
  Query: \{query\}\\
  Model output: \{annotation\}\\
  Correct answer: \{ground\_truth\}\\
}
\end{tcolorbox}
\end{table*}

\begin{table*}
\begin{tcolorbox}[notitle, colback=gray!10, colframe=black, title={Prompts for Dimension Field Refinement (Part 2)},]
\texttt{
\\
- Preliminary intent: \{new\_intent\}\\
- Preliminary boundary: \{new\_boundary\}\\
- Preliminary applicability: \{new\_applicable\}\\
\\
\#\#\# Fusion Analysis Steps (Please follow this reasoning)\\
\\
1. Relationship Determination: Determine the relationship between the new case and historical rules\\
   - Support: New case validates and reinforces historical rules\\
   - Supplement: Fills gaps in scenarios not covered by historical rules\\
   - Correction: Discovers misjudgments or boundaries that are too wide/narrow in historical rules\\
   - Extension: Discovers that historical rules can be generalized to new categories/scenarios\\
2. Essence Extraction: Go beyond surface differences of specific cases to induce general task annotation logic (but don't over-speculate; the extracted logic should serve the task itself)\\
3. Conflict Resolution: If conflicts exist, analyze whether it's a "special handling of boundary cases" or a "fundamental flaw in historical rules", and provide a resolution plan\\
4. Counterexample Testing: Consider what scenarios would produce incorrect judgments if this rule were over-generalized, and tighten the applicable boundary accordingly\\
\\
\#\#\# Output Requirements\\
Note: Please analyze carefully based on the samples themselves. Since some samples may have deviations, please avoid using overly absolute judgment logic.\\
Please avoid academic or abstract vocabulary (such as "mechanism", "paradigm", "dimension", "attribute", etc.), and use plain language to induce general core annotation logic.\\
Please output your complete reasoning process first, then output the final JSON result:\\
Reasoning process: [Your reasoning process]\\
\\
Final JSON format:\\
```json\\
\{\\
  "rule\_intent": "Fused rule intent (reflecting core reasons and generalization principles, but don't deviate from the annotation task itself)",\\
  "decision\_boundary": "Fused decision boundary (clearly define annotation logic)",\\
  "applicable\_boundary": "Fused applicable boundary (define clear scope and exceptions based on sample types, don't over-associate)"\\
\}
```
}
\end{tcolorbox}
\end{table*}

\begin{table*}
\begin{tcolorbox}[notitle, colback=gray!10, colframe=black, title={Prompts for Verifier Prompt Optimization (Part 1)},]
\texttt{\\
System: You are a helpful assistant.\\
\\
User:\\
You are a verification prompt optimization expert. Your task is to analyze current error samples, identify the root causes in the verification prompt that lead to wrong selections, and perform targeted optimization.\\
\\
\#\# Current Verification Prompt (to optimize)\\
\{base\_rules\}\\
\\
\#\# Current Verification Error Samples (core optimization basis)\\
Please carefully analyze the following error samples, understand in which scenarios current verification prompt leads to wrong annotator selections:\\
\\{}
[Error Sample 1:]\\
Query: \{query\}\\
Ground Truth: \{ground\_truth\}\\
Annotator Results:\\
- annotator1: \{annotation\} (label: \{label\})\\
- annotator2: \{annotation\} (label: \{label\})\\
- annotator3: \{annotation\} (label: \{label\})\\
Vote Info: \{vote\_info\}\\
Verification Selected: \{selected\_agent\}\\
Error Analysis: \{error\_reason\}\\
\\{}
[Error Sample 2: ...]\\
\\
\#\# Error Reason Summary\\
\{failure\_reasons\}\\
\\{}
[Optional Auxiliary Reference:]\\
\#\# Auxiliary Reference Information\\
The following dimensions and contrastive samples can help understand error patterns, but optimization should be primarily based on the error samples above:\\
\\
\#\#\#\# Learned Discriminative Dimensions Reference\\{}
[Dimension: \{dim\_name\}] (Evidence count: \{evidence\_count\})\\
- Rule intent: \{rule\_intent\}\\
- Decision boundary: \{decision\_boundary\}\\
- Applicable boundary: \{applicable\_boundary\}\\
}
\end{tcolorbox}
\end{table*}

\begin{table*}
\begin{tcolorbox}[notitle, colback=gray!10, colframe=black, title={Prompts for Verifier Prompt Optimization (Part 2)},]
\texttt{
\\
\#\#\#\# Contrastive Sample Reference (Different judgments for similar Queries)\\{}
[Contrastive Sample 1]\\
Sample A (System judged correct):\\
  Query: \{query\}\\
  System output: \{annotation\}\\
  Correct answer: \{ground\_truth\}\\
Sample B (System judged incorrect):\\
  \{full\_error\_entry\}\\
\\
\#\# Reference Optimization Principles\\
1. Error samples as core: Carefully analyze each error sample, understand why current prompt leads to these errors, optimization should directly solve these problems.\\
2. Reference dimensions and contrastive samples: Dimensions and contrastive samples provide summaries of historical error patterns, can serve as optimization reference, but should not overshadow the main focus.\\
3. Fully integrate related suggestions, don't pile up at the end: Avoid appending "additional notes", "additional clarifications", "supplementary rules" patch paragraphs at the end of the prompt.\\
4. Prioritize reusing existing content: Most optimization should be achieved by extending or rewriting existing prompt content when referencing dimensions and error samples.\\
5. Eliminate contradictions: When modifying or adding content based on dimensions and error samples, must check and synchronously correct other parts that may conflict.\\
\\
\#\# Please strictly output in the following format:\\
Analysis process: [First analyze the root cause of each error sample one by one, then combine with reference information to propose verification prompt optimization plan]\\
Optimized verification prompt: [Fine-tuned complete verification prompt, can be directly used by the verification agent]\\
}
\end{tcolorbox}
\end{table*}

\begin{table*}
\begin{tcolorbox}[notitle, colback=gray!10, colframe=black, title={Prompts for Annotator Prompt Optimization (Part 1)},]
\texttt{
System: You are a helpful assistant.\\
\\
User:\\
You are an annotation Prompt optimization expert. Your task is to analyze current error samples, identify the root causes in the Prompt that lead to annotation errors, and perform targeted optimization.\\
\\
\#\#\# Existing Annotation Prompt (to optimize):\\
\{prompts/guidelines\}\\
\\
\#\#\# Current Annotation Error Samples (Core Optimization Basis)\\
Please carefully analyze the following error samples, understand in which scenarios the current Prompt leads to annotation errors:\\
\\{}
[Error Sample 1:]\\
Query: \{query\}\\
Ground Truth: \{ground\_truth\}\\
Model Output: \{annotation\}\\
Error Analysis: \{error\_reason\}\\
\\{}
[Error Sample 2: ...]\\
\\
\#\#\# Error Reason Summary:\\
\{failure\_reasons\}\\
\\{}
[Optional Auxiliary Reference:]\\
\#\#\# Auxiliary Reference Information\\
The following dimensions and contrastive samples can help you understand error patterns, but optimization should be primarily based on the error samples above:\\
\\
\#\#\#\# Contrastive Sample Reference (Different judgments for similar Queries)\\{}
[Contrastive Sample 1]\\
Sample A (Current Annotator judged correct):\\
  Query: \{query\}\\
  Current Annotator output: \{annotation\}\\
  Correct answer: \{ground\_truth\}\\
\\
Sample B (Current Annotator judged incorrect):\\
  \{full\_error\_entry\}\\
\\
\#\#\#\# Learned Discriminative Dimensions Reference\\
The following dimensions are general discriminative logic induced from historical contrastive samples, for reference only:\\
}
\end{tcolorbox}
\end{table*}

\begin{table*}
\begin{tcolorbox}[notitle, colback=gray!10, colframe=black, title={Prompts for Annotator Prompt Optimization (Part 2)},]
\texttt{\\{}
[Dimension: \{dim\_name\}] (Evidence count: \{evidence\_count\})\\
- Rule intent: \{rule\_intent\}\\
- Decision boundary: \{decision\_boundary\}\\
- Applicable boundary: \{applicable\_boundary\}\\
\\
\#\#\# Reference Optimization Principles\\
1. Error samples as core: Carefully analyze each error sample, understand why current prompt leads to these errors, optimization should directly solve these problems.\\
2. Reference dimensions and contrastive samples: Dimensions and contrastive samples provide summaries of historical error patterns, can serve as optimization reference, but should not overshadow the main focus.\\
3. Fully integrate related suggestions, don't pile up at the end: Avoid appending "additional notes", "additional clarifications", "supplementary rules" patch paragraphs at the end of the prompt.\\
4. Prioritize reusing existing content: Most optimization should be achieved by extending or rewriting existing prompt content when referencing dimensions and error samples.\\
5. Eliminate contradictions: When modifying or adding content based on dimensions and error samples, must check and synchronously correct other parts that may conflict.\\
\\
\#\#\# Please strictly output in the following format:\\
Analysis process: [First analyze the root cause of each error sample one by one, then combine with reference information to propose Prompt optimization suggestions]\\
Optimized Prompt: [Fine-tuned complete Prompt, can be directly used for annotation]\\
}
\end{tcolorbox}
\end{table*}

\end{document}

%% file: methods.tex
\section{Our Proposed ANT-OR Framework}

\subsection{Overview}
\label{sec:overview}

As shown in Figure~\ref{fig:framework}, we propose \textbf{ANT-OR}, an automated \textbf{AN}notation framework \textbf{T}hrough multi-agent \textbf{O}ptimization with agent-level cont\textbf{R}astive knowledge. 
Designed for complex real-world annotation tasks, ANT-OR optimizes agents using contrastive evidence from negative and similar positive samples, uncovering the decision criteria that distinguish confusable categories in boundary cases.

\subsection{Contrastive Knowledge Mining}
\label{sec:knowledge_base_construction}

ANT-OR equips each agent with a structured knowledge base consisting of validated samples and discriminative dimensions, allowing agents to accumulate, in a continual manner, both concrete annotation examples and decision-making knowledge across optimization iterations.

\paragraph{Contrastive Sample Pool.}
We construct a positive sample $s^+=(q, \hat{y}^+)$ with correct output $\hat{y}^+$ on a given query $q$ and $s^-=(q, \hat{y}^-)$ as the negative sample. (In our proposed ANT-A settings, a prediction is considered positive if it matches at least two human labels out of five.) For each agent $a$, we maintain a positive pool $\mathcal{S}_a^+$ and a negative pool $\mathcal{S}_a^-$ of verified instances. During annotation, instances from both pools are sampled into the agent's prompt as instance-level references. Each instance s has a quality score $w$ that represents its sampling priority and is initialized to 1.

\paragraph{Discriminative Dimension Pool.} 
Beyond utilizing instance-level references, agents further extract rule-level experiences. When an agent makes a wrong prediction $\hat{y}^-$ on a given query $q$, we first construct the error sample $s^-$. Then, we retrieve two semantically similar correct samples $s_1^+, s_2^+ \in \mathcal{S}_a^+$ to form a contrastive tuple $(s_1^+, s_2^+, s^-)$ and an LLM is prompted to summarize \textit{discriminative dimension} $d$ from these confusing cases:
\begin{equation}
d = \mathrm{LLM}^{\text{Mine}}(s_1^+, s_2^+, s^-),
\end{equation}
The discriminative dimension is formulated as a structured textual unit, containing category-specific decision criteria distilled from contrastive cases and the applicable scenarios in which these criteria are expected to guide annotation decisions.

To avoid redundant dimensions, for each newly extracted discriminative dimension $d$, we retrieve semantically similar dimensions from the discriminative dimension pool $\mathcal{D}$. When a matching dimension $d_{\text{old}}$ is identified, we apply an LLM-based refinement step to fuse $d$ with $d_{\text{old}}$ before adding the resulting dimension back to $\mathcal{D}$:
\begin{equation}
d := \mathrm{LLM}^{\text{Refine}}(d_{\text{old}}, d).
\end{equation}
Each dimension is assigned a support count, which records the number of times it has been involved in fusion. The support count serves as an estimate of how many training samples with inference errors share similar judgment patterns. To suppress spurious patterns, we retain only dimensions whose support count is no less than 2. We maintain a bidirectional many-to-many mapping between dimensions and source samples for traceability.

\subsection{Knowledge-Augmented Reasoning}
\label{sec:mas}

\paragraph{Annotator--Verifier Architecture.}
Our ANT-OR adopts a hierarchical annotator--verifier architecture~\cite{hegazy2025mafa,ahtisham2025orchestration}: three annotators independently produce labels and rationales using separately optimized prompts, and a verifier then compares their justifications to determine the final label. This design preserves proposal diversity while centralizing conflict resolution, avoiding the costly multi-round deliberation of debate-based methods~\cite{du2024agentdebate} and the plan-induced error propagation of planner--worker alternatives~\cite{wang2023plan} (see Appendix~\ref{sec:architecture_comparison}).

\paragraph{Quality-Aware Sample Retrieval.}
Given a query $q$ and a sample $s_i$ from the contrastive sample pool, we compute the retrieval score as
\begin{equation}
\text{Score}(q, s_i) = \text{sim}\bigl(E(q), E(s_i)\bigr) \cdot w_i,
\end{equation}
where $E(\cdot)$ denotes the embedding function, $\text{sim}(\cdot,\cdot)$ denotes cosine similarity. After each retrieval, quality score $w_i$  corresponding to $s_i$ is updated according to the correctness of the resulting prediction for future retrievals:
\begin{equation}
w_i \leftarrow w_i + \Delta w_i,
\end{equation}
where
\begin{equation}
\Delta w_i = \alpha \cdot
\begin{cases}
1-\rho, & \text{if correct}, \\
-\rho, & \text{if incorrect},
\end{cases}
\end{equation}
with $\rho$ denoting the current global accuracy and $\alpha$ the base step size. This update increases the priority of samples that contribute to correct predictions while penalizing those leading to incorrect predictions. As a result, retrieval is guided not only by semantic relevance but also by the historical reliability of stored examples.

\paragraph{Dimension--Sample Coupled Injection.}
We select the top-$K$ highest-scoring samples, and discriminative dimensions associated with them. ANT-OR injects both instance-level and rule-level into prompts, providing both concrete evidence and abstract decision criteria, while existing methods typically inject either examples or rules alone.

\subsection{Attribution-Guided Optimization}
\label{sec:optimization}

ANT-OR performs prompt optimization at the agent level. It first attributes each error to responsible agents and then refine their prompt accordingly.

\paragraph{Error Attribution.}
After feeding a query into the multi-agent system and obtaining the output, error attribution is triggered when the result fails to match at least two human-annotated references out of five. The current query $q$, the agent interaction history $\mathcal{H}$, and the majority label $g$ are fed into an error attribution model for detailed error analysis, which identifies the agents responsible for the labeling error along with an analysis of the error causes, such as intrinsic mistakes or misleading information from upstream models:

\begin{equation}
    \mathcal{M}_{\text{attr}}: (q, \mathcal{H}, g) \rightarrow \{(a_i, r_i)\}_{i=1}^{n}
\end{equation}

where $\mathcal{M}_{\text{attr}}$ is implemented by the same backbone LLM as annotation agent with a dedicated attribution prompt. $a_i$ is the $i$-th responsible agent, $r_i$ is the corresponding error reason for agent $a_i$, and $n$ is the total number of responsible agents identified for the current error.

\paragraph{Agent-Level Prompt Optimization.}
For each flagged agent $a$, we optimize its prompt $p_a$ using error samples with attributed failure reasons $r$, most relevant positive sample $s^+$, and discriminative dimensions $d$  associated with $s^+$:
\begin{equation}
p_a^{\text{opt}} =
\mathrm{LLM}^{\mathrm{Opt}}
\left(
p_a;
\{(q^-, r)\},
s^+,
d
\right).
\end{equation}
This combines concrete failures, contrastive support, and abstract decision rules in a single refinement step, enabling more targeted prompt updates than optimization solely on error samples. After updating the prompts, we re-evaluate the optimized multi-agent system on the current batch and retain the updated prompts only if batch-level performance improves; otherwise, we roll back to the previous prompts to ensure robust updates.

\paragraph{Iterative Knowledge Base Expansion.}
Optimization proceeds over multiple epochs through an iterative cycle of evaluation, knowledge-base update, and prompt refinement. After each iteration, verified prediction outcomes are added to the contrastive sample pools, $\mathcal{S}_a^+$ and $\mathcal{S}_a^-$, expanding the knowledge base available for subsequent retrieval and contrastive mining. In this way, ANT-OR establishes a self-reinforcing optimization loop, where improved prompts lead to higher-quality annotations, and the enriched knowledge base further supports more informative error attribution and more effective prompt optimization.